\documentclass[10pt]{article} % For LaTeX2e
\usepackage[preprint]{tmlr}   % de-anonymised: removes "Under review" and TMLR mentions

\usepackage{amsmath,amsfonts,bm}

\def\eqref#1{equation~\ref{#1}}
\def\1{\bm{1}}

\DeclareMathAlphabet{\mathsfit}{\encodingdefault}{\sfdefault}{m}{sl}
\SetMathAlphabet{\mathsfit}{bold}{\encodingdefault}{\sfdefault}{bx}{n}

\usepackage[hidelinks]{hyperref}
\usepackage{url}
\usepackage{graphicx}
\usepackage{booktabs}
\usepackage{multirow}

\title{When Does Geometry Emerge from Memorization in Transformers?}

\author{\name RamReddy Nishanth Reddy \email nishanthreddy.r24@iiits.in \\
        \addr Indian Institute of Information Technology Sri City, India
        \AND
        \name Sambhav Manohar \email sambhav.m24@iiits.in \\
        \addr Indian Institute of Information Technology Sri City, India
        \AND
        \name Anish Chand Turlapaty \email anish.turlapaty@iiits.in \\
        \addr Indian Institute of Information Technology Sri City, India}

\begin{document}
\maketitle

\maketitle

\begin{abstract}
Transformer models often show structured internal representations on relational tasks, which are often interpreted as geometric organization. Prior work documents such structure via visualization or performance-based analyses, but does not isolate whether perfect memorization alone yields geometric representations. Here, we conduct a controlled study using synthetic relational worlds defined by canonical graph topologies, explicitly training Transformers to perfectly memorize relational structure without imposing constraints that favor or discourage geometry, and examining whether geometric representations arise as a consequence.

Across chains, cycles, regular graphs, and star graphs, models achieve perfect memorization accuracy while internal embeddings do not systematically preserve either global distances or local neighborhoods, indicating reliance on non-geometric, index-based representations. By probing embeddings against shortest-path distance using rank consistency and neighborhood preservation metrics, we show that memorization alone places no requirement on metric organization. Recoverable geometric structure emerges only when the task objective, together with the relational topology, sufficiently constrains node interchangeability, reducing the space of symmetry-equivalent memorization solutions.

Our results show that perfect memorization does not imply emergent geometric structure, and characterize the conditions under which structure arises in learned embeddings.
\end{abstract}

\section{Introduction}

Transformer models are frequently observed to exhibit structured internal representations, such as linear arrangements or clustered layouts, which are often interpreted as evidence that the model has learned a geometric representation of underlying relations \citep{mikolov2013linguistic, hewitt2019structural}. Such observations have motivated the view that geometric organization may emerge naturally during training, and are commonly taken as evidence of relational structure encoded in neural representations \citep{petroni2019language}.

At the same time, achieving perfect accuracy on a relational task does not require preserving distances, neighborhoods, or any notion of metric structure. Prior work has shown that neural models can achieve perfect memorization through arbitrary internal coding schemes, such as index-based or identity-based representations, without analyzing whether such solutions induce any coherent relational or geometric structure \citep{zhang2017understanding, arpit2017closer}. Transformer architectures in particular admit such associative or lookup-style solutions \citep{geva2021transformer}, making high task performance alone insufficient to diagnose whether learned representations are genuinely geometric.

This ambiguity complicates the interpretation of probe-based geometric analyses of learned representations: apparent structure may arise even when representations are not uniquely identifiable or causally required for task performance \citep{hewitt2021conditional, elazar2021amnesic}. In particular, low-dimensional visualizations from techniques such as UMAP can appear highly structured despite substantial distortions in the underlying metric, and therefore cannot, by themselves, establish meaningful geometric organization \citep{mcinnes2018umap}. Accordingly, we adopt an operational definition of geometric organization based on consistency between distances in representation space and shortest-path distances in the underlying relational graph.

We study the relationship between memorization, internal geometric organization, and representational stability in a controlled setting. Transformer models are trained from scratch on synthetic relational graph worlds under objectives that permit perfect memorization without requiring generalization, allowing us to isolate when geometric structure is forced by the task rather than incidental. We show that memorization alone does not reliably induce geometric organization, and that stable geometry emerges only when the task objective and relational topology sufficiently constrain node interchangeability, reducing the space of symmetry-equivalent memorization solutions.

\section{Methodology}

\subsection{Objective and Scope}

We investigate whether perfect memorization of relational structure necessarily induces a geometric internal representation. To isolate this question, we study Transformer models trained under a pure memorization regime, where the learning objective can be satisfied without compositional reasoning, generalization, or abstraction. Our analysis focuses on the geometry of internal representations rather than task performance alone.

All experiments are conducted in closed synthetic environments, allowing us to disentangle the effects of topology, model capacity, and embedding dimensionality without confounds from linguistic priors or pretraining.

\subsection{Why Graphs Rather Than Lexical Data}

We analyze synthetic graph-structured data rather than lexical or natural language inputs to achieve precise experimental control. Lexical data introduce confounding factors such as semantics, frequency effects, distributional similarity, and pretraining priors, which obscure the relationship between memorization and internal representation.

Graphs provide an explicit relational structure with a well-defined notion of distance, symmetry, and topology. This allows direct comparison between internal embedding geometry and ground-truth relational distances, enabling us to attribute observed representational effects to structural properties of the task rather than linguistic artifacts.

\subsection{Relational Graph Worlds}

Each experiment is defined by a graph $G = (V, E)$, where nodes represent discrete entities and edges represent valid successor relations. Given a node $u \in V$, the model is trained to predict one of its successors $v \in \mathcal{N}(u)$, where $\mathcal{N}(u)$ denotes the set of valid successors of $u$. For chain, cycle, star, and 3-regular graphs, successor relations are directed, ensuring that successor prediction is unambiguous and reducing additional symmetry-based degeneracies; for Watts--Strogatz graphs, edges are undirected and represented using bidirectional successor relations. To confirm that this choice of edge directionality is not the primary driver of our results, we also run an ablation where all graph families are trained using bidirectional successor relations; these results are detailed in Appendix~\ref{app:direction_ablation}.

Throughout this work, we use \emph{symmetry} to refer specifically to automorphisms of the underlying graph. Graph automorphisms determine when nodes are structurally interchangeable under relabeling. However, node interchangeability under the training objective may also arise from task-induced equivalence, even in graphs with limited formal symmetry. In both cases, large equivalence classes reduce the identifiability of metric structure, permitting multiple functionally correct but geometrically distinct representations. We consider several canonical graph families, each selected to probe a distinct structural property relevant to the emergence of geometric representations.We report these equivalence classes explicitly for each family, via automorphism group size, orbit count, and induced target entropy, in Appendix~\ref{app:automorphisms}.

\begin{itemize}
    \item \textbf{Chain.} A directed path with a unique successor per node. Although the graph has a clear linear ordering, the successor-prediction objective does not require the model to represent this ordering explicitly, allowing multiple functionally equivalent encodings that need not preserve metric structure.
    \item \textbf{Cycle.} A closed loop with uniform degree and rotational symmetry, making all nodes structurally interchangeable and globally ambiguous.
    \item \textbf{Star.}  A hub-and-spoke topology with one central node and many structurally identical leaves, yielding shallow and degenerate graph distances.
    \item \textbf{3-Regular Graph.} A graph where each node has degree three. The topology is locally uniform but does not impose a unique global ordering, reducing symmetry without introducing privileged nodes.
    \item \textbf{Watts--Strogatz Small-World Graph.} A graph combining local regularity with randomly rewired long-range connections, introducing multiple length scales and partial symmetry breaking.
\end{itemize}

Unless otherwise stated, graphs have $|V|=60$ nodes; we also test $|V|\in{16,150}$ to assess finite-size and scaling effects.
\subsection{Model Inputs and Graph Encoding}
We train Transformers on a graph memorization task where the objective is to predict
the immediate successor of each node in a relational structure. The graph is not
explicitly represented as an adjacency matrix or edge list; instead, graph topology is
encoded implicitly through the training task itself.

\subsubsection{Single-Token Inputs (Successor Prediction)}
For each node $u \in V$, we construct a single input token $i_u = u$ (the node's
integer ID, in the range $[0, n)$) and a target label $y_u \in N(u)$, where $N(u)$
denotes the set of valid successors of $u$ in the directed graph (Section~2.3).

The model receives input
\begin{align}
\mathbf{x} &= \text{embed}(u) \in \mathbb{R}^{1 \times d}
\end{align}
where $\text{embed}(u)$ is the learned embedding of node $u$. The model processes this
single token through all layers and outputs logits $\boldsymbol{\ell}_u \in
\mathbb{R}^{n}$ over the vocabulary $[0, n)$, trained with cross-entropy loss against
$y_u$ as described in Section~2.5.

The graph topology is not provided as structured input; rather, it constrains the
training objective. The model must memorize the mapping $u \to N(u)$ by learning
internal representations that capture this structure implicitly.

\subsubsection{Multi-Token Inputs (Random Walk Prediction)}
To test whether geometry emerges even when the model processes longer sequences, we
also train on random walks. A walk of length $L$ is a sequence $(u_1, u_2, \ldots,
u_L)$ where $u_{i+1} \in N(u_i)$ for all $i < L$. For the experiments presented in this paper, we fix the walk length to $L=8$.

The model receives input
\begin{align}
\mathbf{x} = \big(\text{embed}(u_1), \text{embed}(u_2), \ldots,
\text{embed}(u_{L-1})\big) \in \mathbb{R}^{(L-1) \times d}
\end{align}
and is trained, at each position, to predict the successor of the corresponding node
under the same cross-entropy objective used for single-token inputs, applied
position-wise across the walk.

Again, the graph structure is entirely implicit: encoded only through the sequential
consistency constraint that each predicted node must lie in $N(u_i)$, the set of
valid successors of the previous node in the walk.

\subsubsection{Graph Information Encoding}
The key design choice is that the graph is not provided as explicit input: there is no
adjacency matrix, no edge features, and no position embeddings encoding distance. The
node's integer ID is the only input feature, and graph topology is encoded
\emph{implicitly} in the training labels through $N(u)$ for each node. The model must
therefore discover representations that distinguish nodes and capture their relational
structure purely through memorization of the successor mapping, rather than from any
structural information supplied directly at the input.

This design allows us to isolate whether memorization \emph{alone} without explicit
geometric inductive biases yields organized internal representations. If the model
learns latent geometries, those geometries emerge as a consequence of the learning
objective, not as a prior encoded in the input.

\subsection{Model Architecture and Training Regime}

We use a decoder-only Transformer trained from scratch, without pretraining or auxiliary objectives. Each node $u$ is mapped to a unique discrete token. For the single-token successor-prediction task (Section~2.4.1), the model receives a single node token and predicts a distribution over node tokens for valid successors; the training objective therefore does not engage sequence-level dependencies or multi-step autoregressive structure. We additionally evaluate a multi-token random-walk variant of the task (Section~2.4.2), in which the model processes a sequence of node tokens and predicts a successor at each position, engaging autoregressive context directly. To verify that our findings are not artifacts of self-attention mechanisms or the decoder-only architecture, we systematically ablate the model class. We replicate our core experiments using an encoder-decoder framework, non-attention residual networks, multi-layer perceptrons (MLPs), and a simple embedding-plus-linear-head model, all of which yield consistent conclusions regarding geometric emergence (Appendix~\ref{app:architecture_independence}).

For both tasks, models are trained with standard cross-entropy (CE) loss for next-node prediction. This objective utilizes a hard target for the specific successor observed in the sampled sequence, permitting the model to assign high probability to a single valid transition without explicitly enforcing uniformity across all possible neighbors. For the single-token task, we also ran a small-scale comparison using a $1/k$ uniform soft-target distribution over all valid neighbors (where $k = |N(u)|$) and observed no qualitative change in the resulting internal geometry. We report hard-target CE-based results throughout, as it seamlessly unifies the single-token and autoregressive multi-hop settings under a standard, least-constraining objective.

Training uses a fixed number of epochs sufficient to reach stable convergence and near-perfect set accuracy ($\geq 99.9\%$) across all graph families. In deterministic topologies, such as chains and cycles, the cross-entropy loss approaches zero. In stochastic topologies with multiple valid successors, the loss stabilizes at a non-zero value determined by the underlying successor distribution, approaching the theoretical task-induced conditional entropy bound.

\subsection{Embedding Dimension and Graph Size Sweeps}

To disentangle geometric effects from representational capacity, we vary both embedding dimensionality and graph size.

Embedding dimension $d$ is varied (e.g., $32, 64, 128, 256$), testing whether geometric organization emerges from increased capacity. Graph size $|V|$ is varied across $16, 60,$ and $150$ nodes, probing whether larger relational worlds constrain representations more strongly or permit non-geometric memorization strategies to persist.

All other architectural and optimization settings are held fixed across these sweeps.

\subsection{Verification of Memorization}

To verify that models successfully memorize the relational structure, we compute
\emph{set accuracy}. For each node $u$, a prediction is counted as correct if the
model's argmax prediction lies in $\mathcal{N}(u)$. For the multi-hop task
(Section~2.4.2), set accuracy is computed identically at each position of the walk and
averaged across positions. Across all reported settings, models achieve set accuracy
above $99.9\%$, ensuring that differences in internal representations are not
attributable to task failure. We additionally verify that models memorize the full
successor neighborhood rather than a single edge (Appendix~\ref{app:neighborhood_recall}).

\subsection{Geometric Probing of Internal Representations}

To assess whether memorization corresponds to geometric organization, we analyze frozen hidden representations at every layer. Let $d_G(u,v)$ denote the shortest-path distance between nodes $u$ and $v$ in the graph, and let
$d_E(u,v) = \lVert e_u - e_v \rVert_{\!2}$ denote the Euclidean distance between their embedding vectors. Graph distances are computed using shortest paths; node pairs that are unreachable under the graph are excluded from all distance-based probes.

\paragraph{Spearman Rank Correlation.}
\begin{equation}
\rho = \mathrm{Spearman}\big(d_G(u,v), d_E(u,v)\big)
\end{equation}
This measures whether embedding distances preserve the ordering of graph distances. High values indicate that closer nodes in the graph tend to be closer in the embedding space, even if absolute scaling is distorted.

\paragraph{Coefficient of Determination ($R^2$).}
\begin{equation}
R^2 = 1 - \frac{\sum (d_E - \hat{d}_E)^2}{\sum (d_E - \bar{d}_E)^2},
\quad \hat{d}_E = a\, d_G + b
\end{equation}
Here, $\hat d_E$ denotes the \emph{least-squares linear fit} of embedding distances $d_E(u,v)$ to graph shortest-path distances $d_G(u,v)$ over all node pairs $(u,v)$, and $\bar d_E$ denotes the mean embedding distance. This measures how well graph distances linearly predict embedding distances; high values indicate proportional alignment consistent with a global metric embedding.

\paragraph{Metric Distortion.}
\begin{equation}
\mathrm{Distortion} =
\frac{\max_{u \neq v} \frac{d_E(u,v)}{d_G(u,v)}}
{\min_{u \neq v} \frac{d_E(u,v)}{d_G(u,v)}}
\end{equation}
This quantifies the uniformity of distance scaling in a worst-case sense. Low distortion indicates the absence of extreme pairwise scaling deviations, while high distortion reflects uneven stretching or compression.

\paragraph{Expansion Variance.}
\begin{equation}
\mathrm{Var}\!\left(\frac{d_E(u,v)}{d_G(u,v)}\right)
\end{equation}

Expansion variance measures how uniformly graph distances are scaled in embedding space. If a coherent global metric is recovered, distances at different graph scales are expanded by similar factors, resulting in relatively low variance. Topologies that exhibit asymmetry, boundary effects, or anisotropy can exhibit non-uniform scaling even when geometry is well formed. In contrast, graphs with degenerate distance structure can produce artificially low variance because many node pairs share identical graph distances.

\paragraph{Local Isometry (1-Hop Stretch).}
\begin{equation}
\frac{1}{|V|} \sum_{u \in V} \frac{1}{|\mathcal{N}(u)|}
\sum_{v \in \mathcal{N}(u)} d_E(u,v)
\end{equation}
Local isometry measures how consistently immediate graph neighborhoods are represented in embedding space. When geometry is well formed, neighboring nodes are embedded at comparable distances, indicating stable local structure. Expected values depend on topology, with shallow neighborhoods yielding smaller values and richer local connectivity producing larger values. This metric is therefore interpreted relative to graph structure rather than as an absolute criterion.

To ensure our structural metrics measure true geometric distortion rather than arbitrary volume scaling of the Transformer's hidden states, we also evaluate scale-invariant companions to these metrics, obtained by taking ratios of raw-scale quantities rather than by rescaling the embeddings themselves. Detailed results isolating this pure geometric signal, alongside additional neighborhood connectivity metrics, are provided in Appendix~\ref{app:scale_free_metrics}.

\subsection{Interpretation of Geometric Outcomes}

A model is considered to exhibit \emph{recoverable geometric organization} if embedding distances reflect graph distances under our probes, as indicated by high Spearman ($\rho$), high $R^2$, controlled distortion, and \emph{topology-consistent} expansion variance. Local isometry is used as a complementary measure of neighborhood coherence and interpreted comparatively across graph families. Conversely, a model is considered non-geometric if it achieves perfect memorization while failing to preserve globally or locally consistent metric structure.

\section{Results}

We evaluate whether geometric structure necessarily emerges from memorization by analyzing internal representations of decoder-only Transformers trained on synthetic relational graphs. Results are reported for graphs with $n = 60$ nodes and embedding dimension $d = 128$, a setting that is large enough to admit non-trivial relational structure while remaining small enough to allow detailed geometric analysis; additional experiments across node counts (Appendix~\ref{app:graph_size}), embedding dimensions (Appendix~\ref{app:dim_sweep}), architecture ablations (Appendix~\ref{app:architecture_independence}), and layer-wise geometric progression (Appendix~\ref{app:layerwise_geometry}) are provided in the Appendix.

\subsection{Perfect Memorization Does Not Imply Geometry}

Despite identical training objectives and perfect memorization accuracy across all graph families, internal representations differ in their geometric properties. Final-layer geometric probe results reveal that memorization alone is insufficient to guarantee the emergence of a coherent metric structure.
\begin{figure}[t]
    \centering
    \includegraphics[width=\textwidth]{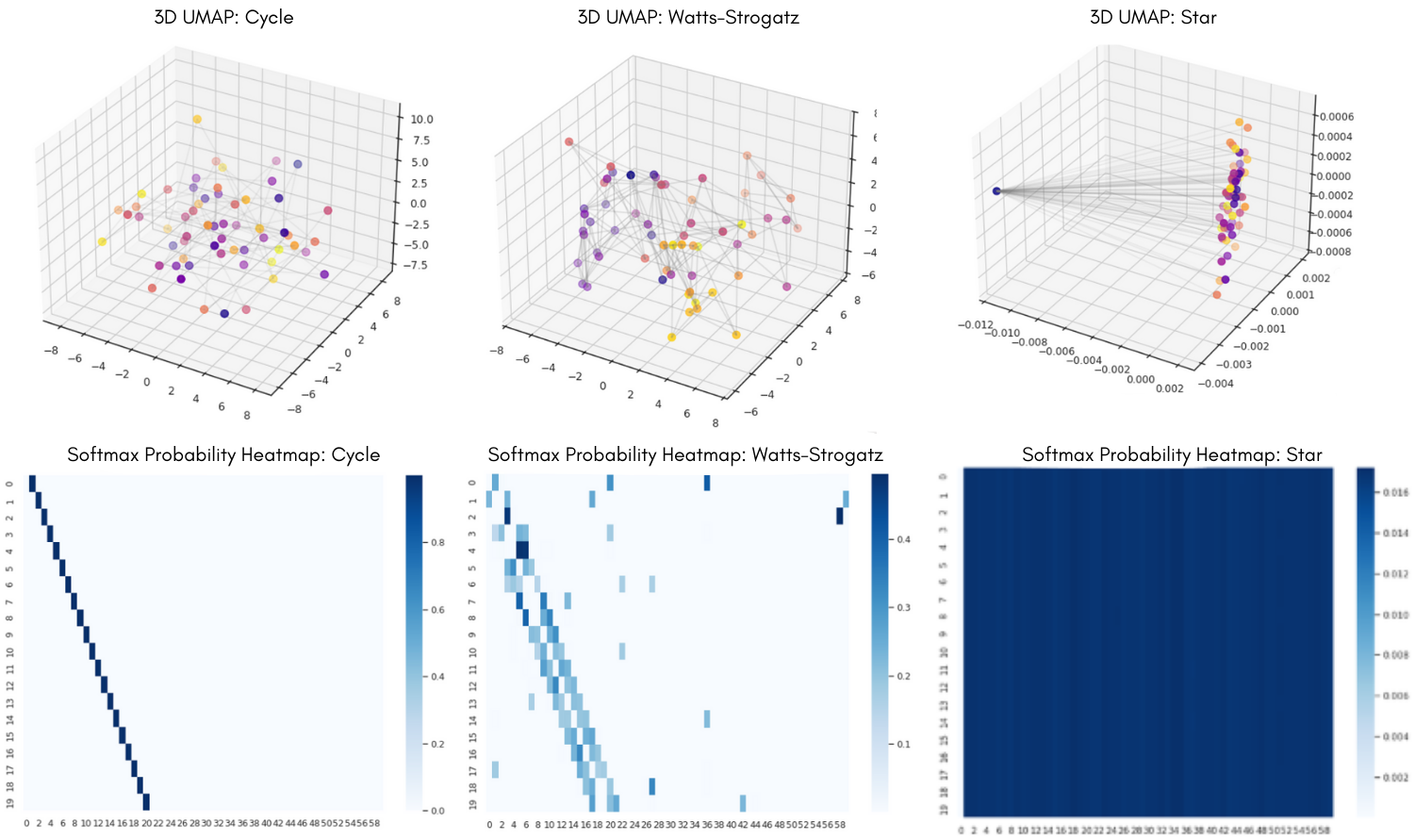}
    \caption{
    \textbf{Geometry vs. memorization across graph topologies.}
    Top: 3D projections of learned node embeddings.
    Bottom: successor probability heatmaps.
    While the model perfectly memorizes local transitions in all cases (bottom),partially recoverable geometric organization emerges only for topologies with sufficient asymmetry (e.g., Watts–Strogatz),
    and collapses under hub–spoke structure (Star).
    }
    \label{fig:geometry_vs_memorization}
\end{figure}
\begin{table*}[t]
\centering
\small
\setlength{\tabcolsep}{8pt}
\begin{tabular}{lccccc}
\toprule
\textbf{Graph} & \textbf{Spearman $\rho$} & \textbf{$R^2$} & \textbf{Distort.} & \textbf{Exp. Var} & \textbf{Loc. Iso} \\
\midrule
Cycle           & 0.038 $\pm$ 0.020 & 0.000 $\pm$ 0.000 & 61.90 $\pm$ 1.25 & 10.68 $\pm$ 0.12 & 16.79 $\pm$ 0.10 \\
Chain           & 0.038 $\pm$ 0.018 & 0.000 $\pm$ 0.000 & 63.65 $\pm$ 3.35 & 10.64 $\pm$ 0.08 & 16.72 $\pm$ 0.06 \\
3-Regular       & 0.326 $\pm$ 0.086 & 0.053 $\pm$ 0.034 & 10.41 $\pm$ 5.61 & 21.87 $\pm$ 1.47 & 15.50 $\pm$ 0.58 \\
Watts--Strogatz & 0.630 $\pm$ 0.034 & 0.428 $\pm$ 0.039 & 8.42  $\pm$ 0.46 & 5.91  $\pm$ 0.32 & 12.83 $\pm$ 0.34 \\
Star            & N/A$^{*}$ & 0.000 $\pm$ 0.000 & 1.54 $\pm$ 0.39 & 0.00 $\pm$ 0.00 & 0.02 $\pm$ 0.01 \\
\bottomrule
\end{tabular}
\caption{Aggregate final-layer geometric probe results across multiple random seeds as well as multiple graph instances ($n=60, d=128$). Results report Mean $\pm$ SD. *Spearman $\rho$ is undefined for Star under the directed convention: only hub--spoke pairs are reachable, so $\mathrm{Var}(d_G)=0$ and $\rho$'s denominator vanishes. $R^2$ is well-defined in this regime  the least-squares fit reduces to the constant predictor $\hat d_E=\bar d_E$, giving $R^2=0$ exactly, which we report rather than omit. Run on 10 seeds, 5 graph instances.}
\label{tab:comprehensive_seeds}
\end{table*}

\subsection{Graphs with High Node Interchangeability Collapse to Non-Geometric Memorization}
Quantitative probe results for all graph families are reported in Table~\ref{tab:comprehensive_seeds}.
\paragraph{Chain and Cycle.}
For both the chain and cycle graphs, Spearman rank correlation between embedding distances and graph distances remains near zero, and distortion values are extremely high.Elevated expansion variance in the absence of rank alignment indicates non-uniform scaling unrelated to graph distance, consistent with projection artifacts rather than meaningful geometry.

As shown in Figure~\ref{fig:geometry_vs_memorization} (column 1), 3D projections of cycle form scattered point clouds without global ordering. Nevertheless, the corresponding softmax probability distributions are sharply peaked, confirming that successor prediction is memorized precisely. These results indicate that relational worlds with high node interchangeability permit index-based or lookup-style representations that do not require geometry.

\subsection{Partial symmetry-breaking: fragile / unstable geometry}

\paragraph{3-Regular Graph.}
The 3-regular graph introduces limited asymmetry while preserving uniform degree. Embeddings show moderate rank correlation, indicating partial alignment between embedding and graph distances. While distortion shows no extreme failures, expansion variance and local isometry are inconsistent with the global structure supported by this topology, reflecting non-uniform scaling and unstable neighborhood structure. As a result, global ordering remains weak. These results indicate that limited asymmetry permits weak but incomplete geometric alignment, insufficient to support a stable global metric embedding.

\subsection{Stronger symmetry-breaking: recoverable geometry}

\paragraph{Watts--Strogatz Graph.}
The Watts--Strogatz small-world graph reduces global symmetry while preserving local connectivity. Embeddings achieve higher rank correlation and substantially higher $R^2$, indicating stronger recovery of global distance ordering, with moderate distortion and relatively low, topology-consistent expansion variance. Embeddings in Figure~\ref{fig:geometry_vs_memorization} (column 2) form a coherent low-dimensional structure in which graph distances are partially recoverable, and successor probabilities assign mass to multiple valid successors due to the presence of multiple admissible paths. Together, these results show that stronger symmetry breaking constrains memorization sufficiently for geometry to become recoverable under our probes.

\subsection{Star Graph as a Structural Control}

The star graph is a control case in which geometric structure is degenerate and imposed by the graph topology. Rank-based metrics are undefined due to the collapse of graph distances, while distortion, local isometry, and expansion variance take trivially non-informative values due to distance degeneracy.

As shown in Figure~\ref{fig:geometry_vs_memorization} (column 3), the hub--spoke topology enforces a simple geometric separation between the hub and spokes. This confirms that the probes distinguish geometry imposed by topology from geometric organization that arises under relational constraints.

\subsection{Geometric Emergence Under Multi-Hop Prediction}
The results in Table ~\ref{tab:comprehensive_seeds} are obtained  under the single-token successor-prediction task
(Section~2.4.1), in which the model receives no sequential context. To test whether
recoverable geometric organization depends on this restricted input structure, we
repeat our probing analysis under the multi-token random-walk task
(Section~2.4.2), in which the model predicts a successor at each position of a
sampled walk and therefore has access to autoregressive context across prior nodes.
All other settings ($n=60$, $d=128$) are held fixed, and models again achieve
near-perfect set accuracy under this task. Aggregate final-layer probe results,are reported in
Table~\ref{tab:multihop_geometry}.

Consistent with the single-token results in Table~1, Chain and Cycle remain
non-geometric under the multi-hop task, with near-zero rank correlation and high
distortion. 3-Regular and Watts--Strogatz both show substantially higher rank
correlation and $R^2$ than the symmetric topologies, with Watts--Strogatz again showing
the strongest alignment between embedding and graph distances. As in the single-token
setting, rank-based probes remain undefined for the Star graph due to the collapse of
ground-truth shortest-path distances. These results indicate that the qualitative
pattern relating topological symmetry to geometric recoverability is not an artifact of
the single-token input structure, and persists when the model is instead trained under
an autoregressive, multi-step formulation of the same relational task.

\begin{table*}[t]
\centering
\small
\setlength{\tabcolsep}{8pt}
\begin{tabular}{lccccc}
\toprule
\textbf{Graph} & Spearman $\rho$ & $R^2$ & Distort. & Exp. Var & Loc. Iso \\
\midrule
Cycle           & $0.012 \pm 0.041$ & $0.002 \pm 0.002$ & $15.854 \pm 0.376$ & $9.294 \pm 0.535$ & $15.679 \pm 0.491$ \\
Chain           & $0.022 \pm 0.022$ & $0.001 \pm 0.001$ & $25.243 \pm 0.899$ & $9.181 \pm 0.437$ & $15.474 \pm 0.378$ \\
3-Regular       & $0.384 \pm 0.033$ & $0.223 \pm 0.025$ & $4.781 \pm 0.242$ & $7.335 \pm 0.284$ & $15.080 \pm 0.304$ \\
Watts--Strogatz & $0.524 \pm 0.027$ & $0.340 \pm 0.030$ & $4.662 \pm 0.316$ & $6.509 \pm 0.225$ & $13.136 \pm 0.283$ \\
Star            & N/A$^{*}$ & N/A$^{*}$ & $4.423 \pm 1.120$ & $6.544 \pm 0.728$ & $15.059 \pm 0.729$ \\
\bottomrule
\end{tabular}
\caption{Geometry emergence under the multi-hop (8-hop walk) sequence task ($n=60, d=128$). Each family is trained and probed under the only edge direction that admits a length-8 walk on it: chain and cycle directed, 3-regular, Watts--Strogatz and star bidirectional. Results show that the topology-dependent pattern of geometric emergence persists under multi-hop path prediction, not just single-step successor prediction. Aggregated across multiple random seeds and, for 3-regular and Watts--Strogatz, multiple graph instances. $^{*}$Spearman $\rho$ and $R^2$ undefined for Star: near-degenerate with only two distance classes.Run on 10 seeds; 5 graph instances.}
\label{tab:multihop_geometry}
\end{table*}

\subsection{Summary}

Across all experiments, we observe a consistent pattern: memorization is compatible with multiple representational strategies, only some of which are geometric. Relational worlds with large symmetry-equivalence classes permit non-geometric solutions, partial symmetry-breaking yields fragile geometry, and sufficiently constrained topologies enable stable recoverable geometric organization. Together, these results show that geometric structure is not an automatic consequence of memorization, but instead depends critically on the topological and symmetry constraints of the underlying relational environment.
\section{Discussion}

\subsection{Memorization Permits Multiple Representational Realizations}
Our experiments show that perfect memorization does not uniquely determine the geometric organization of internal representations. Across all graph families, models achieve near-perfect set accuracy and approach their task-dependent entropy lower bound, yet exhibit markedly different geometric properties when probed. This reflects that the training objective admits multiple internal realizations that are functionally equivalent for memorization.

Whether geometric structure becomes recoverable depends on how relational constraints are jointly imposed by topology and task. Topology-induced constraints arise from structural asymmetries in the graph itself, while task-induced constraints arise from the successor-prediction objective, which may render distinct nodes functionally interchangeable. Graphs such as Chains and Cycles possess well-defined topological structure, but the task objective does not expose this structure, permitting non-geometric memorization solutions. The 3-regular graph introduces partial asymmetry, weakly restricting interchangeability and inducing fragile, locally coherent geometry that does not stabilize into a global metric. Conversely, the Star graph imposes strong topological asymmetry, but the task remains degenerate due to collapsed distance structure, yielding only trivial, topology-imposed geometry. In contrast, the Watts–Strogatz graph introduces persistent structural heterogeneity that, when coupled with the successor-prediction objective, restricts the space of admissible memorization solutions enough for graph distances to become partially recoverable from embedding distances under our probes.

\subsection{Symmetry Suppresses Metric Identifiability}
Here we formalize symmetry in terms of graph automorphisms, which determine when nodes are interchangeable under the training objective. Graph families such as cycles admit large automorphism groups, while others such as chains exhibit task-induced interchangeability under the successor objective despite limited formal symmetry.Appendix~\ref{app:automorphisms} quantifies this directly: $|Aut(G)|$, orbit count, and target entropy alongside Spearman $\rho$ for each family, and a Watts--Strogatz rewiring sweep that isolates formal automorphism-breaking from functional (task-induced) symmetry breaking — rewiring collapses $|Aut(G)|$ from $120$ to $1$ while target entropy barely moves, yet rank correlation strengthens accordingly.
From a representational perspective, this implies that multiple embeddings related by permutation or rotation are equally valid solutions to the training objective. As a consequence, there is no unique or preferred metric ordering that must be recovered in embedding space.

This symmetry manifests empirically as near-zero rank correlation between embedding distances and graph distances, along with persistently high distortion. Because successor-preserving relabelings, whether arising from graph automorphisms or from task-induced equivalence, do not affect prediction accuracy, embeddings need not align with graph-theoretic distances.

\subsection{Symmetry-Breaking Constrains Representational Degrees of Freedom}

Graph topologies that introduce asymmetry or irregular connectivity reduce the space of functionally equivalent representations. When nodes differ in their relational roles, automorphism-induced equivalence classes collapse, and representational solutions must respect these distinctions to satisfy the training objective.

In such settings, embedding distances become more strongly coupled to graph distances, resulting in improved rank consistency and reduced distortion. However, this coupling is conditional rather than automatic. Increasing embedding dimension expands the space of admissible representations and may suppress geometric alignment unless relational asymmetry sufficiently constrains non-metric solutions. This explains why geometry does not emerge monotonically with capacity, and why scale alone does not guarantee metric structure.

\subsection{Capacity, Scale, and Robustness Across $n$ and $d$}

Varying the number of nodes and embedding dimension confirms that the observed behaviors are robust to scale. In symmetric graph families, increasing $n$ or $d$ preserves near-perfect memorization while leaving geometric probes unchanged, indicating that scale does not resolve metric non-identifiability. In contrast, for asymmetric graph families, increased capacity stabilizes geometric organization when topological constraints limit alternative realizations.

These findings indicate that geometric structure arises from the interaction between model capacity and relational constraints, rather than from scale or optimization dynamics alone.

\subsection{Boundary Cases and Implications}

The star graph provides a boundary case in which geometric structure is imposed directly by topology rather than resolved by the task. Owing to its hub--spoke organization, geometry is trivial and  unavoidable and rank-based probes are degenerate by construction. This illustrates that geometric organization may be imposed, recoverable, or absent depending on the relational structure.

More broadly, these findings indicate that geometric analyses of learned representations should be interpreted relative to task identifiability, rather than treated as intrinsic properties of the architecture. When the training objective admits multiple functionally equivalent solutions, the presence or absence of geometric structure alone is insufficient to diagnose how relational information is encoded.

\section{Related Work}

\subsection{Geometric Structure in Transformer Representations}

A growing literature reports that Transformer models often exhibit structured or geometric organization in their hidden representations. Early evidence of linear and hierarchical structure in neural embeddings emerged from studies of word representations \citep{mikolov2013distributed, pennington2014glove}, followed by analyses showing that contextual Transformers encode syntactic and semantic relations in linearly recoverable form \citep{hewitt2019structural, jawahar2019bert}. More recent work has studied the geometric structure of transformer representations across layers and tasks \citep{valeriani2023geometry}.

\subsection{Associative Memory, Lookup Tables, and Memorization}

A long-standing view models neural memorization as associative or lookup-based storage of co-occurrences, where embeddings act as arbitrary identifiers and relations are stored in weight matrices \citep{hopfield1982neural}. This abstraction has been widely adopted in analyses of Transformer memory \citep{geva2021transformer}, and is often sufficient to explain behavior on disjoint or atomic facts.

Several works argue that, in the absence of structural constraints, Transformers can implement near-perfect lookup-table solutions \citep{arpit2017closer}. Recent empirical studies suggest that increased model capacity primarily improves memorization fidelity rather than inducing structured representations. Our findings align with this view in symmetric relational worlds, while clarifying the conditions under which associative solutions coexist with or give way to geometric ones.

\subsection{Probing, Identifiability, and Representation Analysis}

Probing methods are commonly used to assess what information is encoded in neural representations \citep{adi2016fine, conneau2018you}. However, multiple works caution that probe success does not imply that a representation intrinsically encodes a property, as probes may exploit incidental correlations or task structure \citep{hewitt2021conditional, ravichander2021probing}. These observations motivate metric, rank, and local-structure probes beyond linear separability \citep{ethayarajh2019contextual}, alongside information-theoretic alternatives \citep{pimentel2020information}. Our results highlight a limitation of geometric probing analyses commonly used in natural language processing. When the training objective admits multiple functionally equivalent representations, observed geometric structure may be incidental rather than identifiable. Probing results should therefore be interpreted relative to the constraints imposed by the task, rather than assumed to reflect learned relational understanding.

Our approach builds on this perspective by evaluating geometric organization using complementary distance-based probes and by analyzing robustness across scales. We connect geometric probing to task identifiability: when multiple representational realizations are functionally equivalent, probes may reveal the absence of a uniquely recoverable geometry.

\subsection{Relational and Graph-Based Learning}

Relational reasoning has traditionally been studied using graph neural networks, which impose explicit inductive biases for locality and message passing \citep{kipf2016semi, hamilton2017inductive}. Transformers, despite lacking such biases, have shown strong empirical performance on graph and relational tasks \citep{dwivedi2023benchmarking}. Several works explore hybrid or geometry-aware architectures that explicitly encourage metric structure \citep{chen2022structure}.

Our work differs from prior work by holding architecture fixed and varying relational topology. This allows us to isolate how symmetry and asymmetry in the task constrain representational geometry, independent of architectural enforcement.

\section{Conclusion}

We investigated whether geometric organization is a necessary consequence of memorization in Transformer models by analyzing internal representations learned on synthetic relational graph worlds. Across a range of graph topologies, node counts, and embedding dimensions, we show that perfect memorization is compatible with multiple representational realizations, only some of which exhibit coherent geometric structure. Our results identify topology induced symmetry breaking in the task as a central factor governing when geometry is recoverable from the training objective. Relational worlds with large equivalence classes arising from graph symmetry or from task-induced interchangeability admit non-geometric memorization strategies even at scale. Importantly, increasing model capacity alone does not guarantee geometric emergence and may instead expand the space of functionally equivalent, non-metric solutions. These findings clarify the distinction between memorization and representation, and caution against interpreting geometric structure as an intrinsic or automatic property of Transformer architectures. More broadly, our work highlights the role of task identifiability in representation analysis and provides a principled framework for evaluating when geometric organization reflects genuine relational structure rather than functionally equivalent alternatives.

\section{Limitations}
Our study focuses on controlled synthetic relational environments designed to isolate the representational consequences of memorization. While this setting enables precise analysis of identifiability and geometric structure, it does not capture the full complexity of natural language or multimodal data, where semantic grounding, redundancy, and compositional generalization may introduce additional constraints on representations.

We evaluate successor prediction over 1- and 8-hop paths across a broad architectural family including linear models, MLPs, encoder-decoders, and 1--8 layer Transformers. The qualitative pattern we report (topology-dependent geometric emergence, and its collapse under the star graph's induced symmetry) remains robust across all these variations, spanning multiple graph scales ($n=16$ to $150$) and distance conventions (directed vs. undirected). All experiments, however, are conducted at a scale far below that of models trained on natural language or large real-world graphs: our largest configuration uses on the order of 1.5--1.7 million parameters. This scale was a deliberate choice, allowing us to compute exact shortest-path ground truth, probe every layer, and aggregate over enough seeds and graph instances to report meaningful variance, but it also means we cannot directly establish how the effects we observe behave as model and graph size grow toward realistic regimes. We regard this as an open question rather than one this study resolves.

Our analysis characterizes representations after training reaches stable, verified convergence (set accuracy $\geq 99.9\%$, loss plateaued); we probe the identifiability of the converged solution space rather than the optimization trajectory. A dedicated training-dynamics study, tracking when and how geometric structure emerges over training, would address a distinct question and remains open for future work.

While our probes are validated against outlier sensitivity and metric conventions, they remain a fixed family of distance-based diagnostics. Regularization schemes, optimizer choices, or entirely different paradigms (e.g., causal interventions or structural classifiers) remain unexplored and could theoretically reveal organization our metrics miss. Thus, our conclusions concern the existence and stability of recoverable geometric structure under these specific probes, rather than asserting an absolute absence of latent organization.

Appendix~\ref{app:automorphisms} links automorphism group size and task-induced target entropy to geometric recoverability, including a rewiring sweep separating formal from functional symmetry-breaking, but this support is empirical rather than a formal theoretical framework relating equivalence-class size to the dimensionality of the admissible solution space; such a framework is an open direction for future work.

\bibliography{main}
\bibliographystyle{tmlr}

\section{Appendix}
\label{sec:appendix}

\appendix
\section{Experimental Setup, Resources, and Reproducibility}
\label{app:repro}

This section documents the experimental setup, hyperparameters, computational resources, and implementation details required to reproduce the results in the paper. The goal is to ensure transparency and reproducibility rather than to optimize for computational efficiency.

\paragraph{Model architecture and size.}
All experiments use a decoder-only Transformer trained from scratch.
The architecture follows a GPT-2--style decoder with 8 Transformer layers, each employing multi-head self-attention with 4 attention heads over a $d$-dimensional embedding space.
Embedding dimensionality is varied across experiments ($d \in \{32,64,128,256\}$), and graph size is varied over $|V|\in\{16,60,150\}$.
Each node in the graph is mapped to a unique discrete token, and the model predicts successor tokens given a single-node input.
As a reference configuration, for $d=128$ and $|V|=60$ (the primary setting reported in the main paper), the model contains approximately $1.5$--$1.7$ million parameters, depending on bias terms and embedding size.

\paragraph{Training procedure and hyperparameters.}
Models are trained using the AdamW optimizer with learning rate $1\times10^{-3}$, weight decay $0.1$, and batch size $32$.
The training objective is cross-entropy loss over successor prediction.
All runs are trained for a fixed $10{,}000$ epochs, which is sufficient to reach stable convergence and near-perfect set accuracy ($\geq 99.9\%$) across all graph families.
We do not employ early stopping, validation-based model selection, or automated hyperparameter search.
Hyperparameter values were chosen based on standard practice and empirically verified to yield stable convergence across graph families and model sizes.
Hyperparameters were not tuned on held-out evaluation data.

\paragraph{Graph construction and data generation.}
Experiments are conducted on synthetic relational graph worlds constructed using standard graph generators. For chain, cycle, star, and 3-regular graphs, successor relations are directed. For Watts--Strogatz graphs, edges are undirected and represented using bidirectional successor relations. Watts--Strogatz graphs are generated with nearest neighbors $k = 4$ and rewiring probability $p = 0.2$. Unless otherwise stated, graphs contain $|V| = 60$ nodes. Shortest-path distances used in geometric probes are computed with respect to the graph definition for each topology, and node pairs that are unreachable under the corresponding graph are excluded from distance-based analyses.

\paragraph{Random seeds and reporting.}
For the primary configuration reported in the main paper ($d=128$, $|V|=60$), all quantitative results are aggregated over $10$ independent training seeds, together with $5$ independent graph instances for the stochastic topologies (3-Regular and Watts--Strogatz), and reported as mean $\pm$ standard deviation.
For all other configurations (embedding-dimension sweeps, graph-size sweeps, and ablations reported in the Appendix), results are aggregated over $5$ independent training seeds for Chain, Cycle, and Star, and over $5$ training seeds $\times$ $3$ independent graph instances for 3-Regular and Watts--Strogatz.
Unless otherwise stated, all reported metrics correspond to the final Transformer layer after convergence.

\paragraph{Computational resources.}
 All experiments were run on a single NVIDIA A6000 GPU. Training time varied substantially depending on graph topology, graph size, embedding dimension, and random seed. Across all experiments  including embedding-dimension sweeps, graph-size sweeps, loss-function ablations, and multiple random seeds  the total computational budget was on the order of several GPU-hours.

\paragraph{Software and implementation.}
Models are implemented in PyTorch using standard Transformer components.
Graph construction and shortest-path computations use NetworkX.
UMAP is used exclusively for qualitative visualization and is not used for quantitative evaluation or for any claims in the paper.
All experiments rely on publicly available libraries, and no external codebases were modified. Code to reproduce all experiments will be 
made available upon acceptance.

\section{Additional Experimental Results}
\label{app:additional_results}

This appendix provides supplementary experimental results supporting the claims in the main paper.
Unless otherwise stated, all values are reported from the \emph{final layer} of the model (Layer~8),
reflecting representations at convergence.

Most tables use the same five geometric probes as in the main paper: Spearman rank correlation and $R^2$,
where higher values indicate stronger geometric alignment with graph shortest-path distance; distortion,
where lower values indicate greater metric uniformity; and expansion variance and local isometry, which
provide complementary diagnostics of low-dimensional structure and neighborhood preservation.
The dimension-sweep, graph-size-sweep, and loss-ablation tables in this appendix additionally report
Exp.\ CV and 1-hop Norm, the scale-invariant companions to Expansion Variance and Local Isometry defined
in Appendix~\ref{app:scale_invariant_metrics}; readers wanting the full derivation and interpretation of
these two columns should consult that section directly.
Several subsections additionally report scale-invariant companions to these probes
(Appendix~\ref{app:scale_free_metrics}), neighborhood-preservation and trustworthiness probes, and
graph-theoretic symmetry statistics such as automorphism group size, orbit count, and target entropy
(Appendix~\ref{app:automorphisms}); these additional metrics are defined at first use in their
respective subsections.

% ----------------------------------------------------
\subsection{Embedding Dimension Sweeps (Cross-Entropy Loss)}
\label{app:dim_sweep}

We examine how embedding dimensionality affects geometric organization when memorization
accuracy is held constant. Graph size is fixed at $n=60$, and all models are trained using
cross-entropy loss. Across all settings, models achieve perfect memorization, allowing us to
isolate representational geometry from task performance.

\paragraph{Low-dimensional regime ($d=32$).}
Results are reported in Table~\ref{tab:geom_n60_d32}.
At low dimensionality, representational capacity is limited, restricting the set of embedding configurations that can achieve perfect memorization. In Chain and Cycle graphs, this constraint does not induce
geometry the successor-prediction task exposes no global ordering, and nodes remain largely
interchangeable. As a result, embeddings fail to reflect graph distances despite the apparent
simplicity of these topologies.

In contrast, the 3-regular graph already exhibits measurable geometric alignment at $d=32$.
Although all nodes have identical degree, higher-order connectivity patterns differ, introducing
weak but genuine asymmetry. Under limited capacity, these differences partially constrain
representations, yielding non-zero rank correlation. However, scaling remains inconsistent,
indicating that this alignment is local rather than globally metric.

Watts--Strogatz graphs show clearer geometric structure even at low dimensionality. Long-range
shortcuts introduce global asymmetry by differentiating nodes according to path centrality and
reachability, allowing embeddings to reflect graph distances even under tight capacity
constraints.

\paragraph{Intermediate dimensionality ($d=64$).}
Results are reported in Table~\ref{tab:geom_n60_d64}.
As dimensionality increases, representational capacity expands. For Chain and Cycle graphs, this
does not improve geometry: the task still permits index-based solutions, and additional dimensions
only increase the space of non-metric representations.

For the 3-regular graph, increased capacity weakens geometric alignment relative to $d=32$.
Though local neighborhoods remain structured, the graph’s near-regularity permits many nodes to be represented by shared functional roles rather than arranged according to global graph distance. The model can
satisfy the task using these alternative encodings, leading to reduced rank consistency.

Watts--Strogatz graphs, by contrast, show no systematic change at this capacity: rank correlation
and $R^2$ remain close to their $d=32$ values, within the range of seed-to-seed variability already
present at that setting. Shortcut-induced asymmetry continues to constrain the representation
regardless of the extra capacity, so alignment neither improves nor collapses at this intermediate
dimension.

\paragraph{High-dimensional regime ($d=256$).}
Results are reported in Table~\ref{tab:geom_n60_d256}.
At high dimensionality, these trends become more pronounced. Chain and Cycle graphs remain
non-geometric, as additional capacity further decouples representations from any implicit
ordering.

In the 3-regular graph, geometric alignment deteriorates further. Without strong global anchors,
the model increasingly exploits high-dimensional freedom to realize symmetry-equivalent
memorization strategies. This results in unstable geometry characterized by reduced rank
alignment, increased distortion, and noticeable variability across random seeds.

Watts--Strogatz graphs, however, maintain stable geometric organization at higher dimensions.
Once shortcut-induced asymmetry establishes a meaningful global structure, additional dimensions
neither disrupt nor substantially enhance geometry. Instead, embeddings saturate into a
consistent configuration that preserves graph distances under our probes.

\paragraph{Summary.}
Embedding dimensionality interacts with graph topology in a graph-specific manner. Highly interchangeable structures such as Chains and Cycles remain non-geometric across all
dimensions. Graphs with weak but higher-order structural asymmetry, such as 3-regular graphs,
exhibit dimension-sensitive and seed-unstable geometry. In contrast, graphs with persistent
multi-scale asymmetry, such as Watts--Strogatz graphs, support stable geometric organization
across a wide range of embedding dimensions. The star graph forms a distinct boundary case
its degenerate hub--spoke distance structure imposes a trivial geometric separation that
remains unchanged across embedding dimensions.
These results highlight that changes in embedding dimension affect geometry through
their interaction with specific structural features, rather than through capacity alone.

% ----------------------------------------------------
\subsection{Loss Function Ablation}
\label{app:loss_ablation}

We next test whether representational collapse is specific to the cross-entropy objective.
We compare against a uniform $1/K$ loss that assigns equal probability mass to all valid successors,
holding graph size ($n=60$) and embedding dimension ($d=128$) fixed.
All comparisons are reported for the \emph{final layer} only, after models achieve stable convergence
and near-perfect set accuracy.
Aggregate geometric probe results are reported in Table~\ref{tab:geom_n60_d128_invk}.

Across all graph families, the same qualitative pattern of which topologies support geometric
structure holds under both loss functions.
Graph families with high node interchangeability (Chain and Cycle) remain non-geometric under both
objectives, exhibiting near-zero rank correlation and extremely high distortion despite perfect
memorization.
This is consistent with representational collapse in these settings reflecting the task admitting
multiple functionally equivalent, non-metric solutions, rather than being specific to either loss
function.

For the 3-regular graph, which introduces partial asymmetry while maintaining uniform degree,
rank correlation and distortion under the uniform $1/K$ objective are similar in magnitude to those
observed under cross-entropy, with substantial variability across seeds in both cases.
This is consistent with partial symmetry supporting a moderate degree of recoverable structure
under either objective, while the underlying geometry remains sensitive to seed-level variation.

In the Watts--Strogatz graph, which strongly breaks global symmetry while preserving structured
local connectivity, geometric organization remains clearly above the level seen for Chain, Cycle,
or the degenerate Star case under both loss functions.
Under the $1/K$ objective, rank correlation and $R^2$ for Watts--Strogatz and 3-Regular are closer
in magnitude to one another than under cross-entropy, where the two families are more clearly
separated.
Given the seed-level variability already observed for 3-Regular, this difference in separation
between the two objectives should be interpreted with caution rather than as evidence that one
objective systematically favors either topology.

Finally, the star graph exhibits degenerate geometric behavior under both objectives.
Rank-based metrics are undefined due to the lack of variance in graph distances, while distortion and
expansion variance reflect topology-imposed constraints rather than learned metric organization.
Local isometry under the uniform $1/K$ objective takes a different value than under cross-entropy,
consistent with the trivial and identical 1-hop neighborhoods imposed by the hub--spoke topology;
this illustrates that purely local probes become ill-conditioned under extreme distance degeneracy,
rather than indicating meaningful geometric structure.

\paragraph{Summary.}
The two loss functions produce broadly similar qualitative outcomes across graph families: Chain and
Cycle remain non-geometric, and Watts--Strogatz and Star retain their characteristic behavior, under
both objectives.
The quantitative differences observed for 3-Regular and Watts--Strogatz are comparable in scale to
the seed-level variability already present within a single objective, so they are not treated as a
systematic effect of the loss function.
Recoverable geometry at convergence is most consistent with being governed by the topological and
symmetry constraints of the underlying graph.
% ----------------------------------------------------

\subsection{Effect of Graph Size}
\label{app:graph_size}

Finally, we evaluate whether representational collapse is a finite-size artifact by varying graph size under cross-entropy training.

\paragraph{Small graphs.}
Results for smaller graphs are reported in Table~\ref{tab:geom_n16_d32}.
At small graph sizes ($n=16$), geometric alignment emerges selectively depending on the degree of relational identifiability imposed by graph topology. Graph families with limited node interchangeability exhibit measurable geometric alignment, while graph families with high interchangeability continue to admit non-geometric memorization solutions (Table~\ref{tab:geom_n16_d32}). In this small-$n$ regime, each relational constraint influences a large fraction of node pairs, making global geometric probes particularly sensitive to weak symmetry-breaking effects.

At small $n$, partial asymmetries introduced by graph topology constrain the space of admissible representations more strongly, as fewer functionally equivalent embeddings satisfy the training objective. This finite identifiability regime amplifies weak topological constraints that would otherwise be diluted at larger scales. As a result, even incomplete relational structure can induce measurable rank alignment between embedding distances and graph distances.

This effect is most evident in the 3-regular graph. Although all nodes share identical degree, higher-order connectivity patterns differ substantially at $n=16$, including variations in cycle participation and neighborhood overlap. These differences reduce node interchangeability and limit the viability of purely index-based memorization strategies. However, this does not translate into reliable geometric alignment: rank correlation, $R^2$, and distortion all carry substantial variability across seeds, with individual seeds ranging from near-zero to moderate alignment, reflecting the absence of a stable, coherent metric embedding rather than a consistently emergent one.

Similarly, the Watts--Strogatz graph exhibits robust geometric alignment even at small scale. Because the topology introduces persistent structural heterogeneity through overlapping local clustering and long-range shortcuts, these relational asymmetries prevent large equivalence classes of interchangeable nodes from forming. The continuous embedding space is mathematically forced to reflect these overlapping topological constraints, establishing a stable global ordering of nodes. As a result, the model cannot rely on localized memorization strategies, leading to strong rank correlation and bounded distortion.

Highly symmetric graph families such as chains and cycles continue to admit non-geometric memorization strategies even at small scale. In these settings, nodes remain functionally interchangeable under the training objective, allowing the model to achieve perfect memorization without encoding relational distances. Accordingly, rank correlation remains near zero and distortion remains high, indicating representational collapse rather than metric structure.

The star graph represents a distinct boundary case. At $n=16$, graph-theoretic distances are degenerate, consisting primarily of hub--spoke relations with no meaningful ordering among spoke nodes. Ratio-based geometric probes such as distortion are therefore ill-conditioned in this regime; individual spoke placements exert disproportionate influence on extreme distance ratios, leading to elevated distortion values. Importantly, this does not reflect a failure to learn task-relevant structure. The model learns a task-aligned geometric separation between the hub and the set of spokes that is sufficient for perfect successor prediction, while distances among spokes remain unconstrained and unidentifiable.

Together, these observations demonstrate that geometric structure does not arise from small scale alone. Rather, small-$n$ regimes amplify the effect of topological constraints, revealing partial geometry when relational asymmetries restrict the solution space, and exposing metric degeneracies when relational structure remains underdetermined.

\paragraph{Large graphs.}
Results for larger graphs are reported in Table~\ref{tab:geom_n150_d384}.
At larger graph sizes ($n=150$), the behavior of geometric probes diverges sharply across graph families, reflecting how each topology interacts with scale under the memorization objective.

For the 3-regular graph, the partial geometric alignment observed at smaller $n$ degrades significantly. Although node degree remains fixed, increasing graph size renders the topology progressively more homogeneous; most nodes occupy statistically similar positions with respect to local neighborhoods and multi-hop connectivity. As $n$ grows, these similarities dominate the task signal, making many nodes functionally indistinguishable from the model’s perspective. The successor prediction objective can therefore be increasingly satisfied by grouping nodes into a small number of role-based equivalence classes rather than organizing them in a globally consistent metric space. This leads to a severe degradation in geometric alignment, characterized by a substantial drop in rank correlation despite perfect memorization.

Chain and cycle graphs exhibit a different failure mode. While these topologies possess a well-defined global metric structure, the successor prediction task does not require the model to preserve long-range distances as graph size increases. As $n$ grows, the majority of training signal is dominated by local transitions, and errors in global ordering become increasingly irrelevant to the loss. Consequently, embeddings need only preserve local adjacency rather than coherent global geometry, resulting in vanishing rank correlation and severe distortion at scale.

In contrast, Watts--Strogatz graphs maintain broadly stable rank-based geometric alignment even at large $n$: Spearman correlation and $R^2$ remain comparable to, or slightly above, their values at smaller graph sizes. Although locally near-regular, this topology introduces persistent structural heterogeneity through long-range shortcuts, and as the graph grows, differences in shortcut participation and local features continue to prevent large equivalence classes of interchangeable nodes from forming, so the model must preserve global distance relationships to satisfy the training objective. The raw (worst-case) distortion statistic, however, increases substantially and becomes highly variable across seeds at this scale, indicating that a small number of outlier node pairs dominate this particular ratio-based probe at large $n$ -- the same pattern of instability we document for ratio-based probes elsewhere in this study, rather than a breakdown of the underlying rank-based alignment.

Finally, the star graph represents a degenerate boundary case: a single hub imposes an explicit geometric separation from the leaves that follows trivially from the topology rather than from learning. After excluding unreachable node pairs, the star graph admits only a single non-trivial distance class, so rank-based geometric probes are undefined by construction and the star graph is excluded from correlation-based analysis.

Together, these results show that increasing graph size does not inherently promote geometric organization. Instead, scale interacts with topology by either diluting or preserving the structural signals required for geometric identifiability. Topologies whose relational constraints remain informative as $n$ grows continue to support recoverable geometry, while those that become functionally homogeneous or locally sufficient under the task objective admit non-geometric memorization solutions.

\paragraph{Summary.}
Across graph families and scales, increasing graph size alone does not induce geometric organization. When topological constraints are weak or homogeneous, perfect memorization is possible without distance-aware representations, leading to probe collapse. In contrast, topologies with persistent structural heterogeneity support recoverable geometry. These results show that geometric structure is governed by topological identifiability rather than graph size or model capacity.

% ====================================================
\subsection{Qualitative Visualization of Learned Representations}
\label{app:visualizations}

To complement the quantitative geometric probes reported above, we provide qualitative
visualizations of learned representations for each graph family.
These visualizations are intended as illustrative diagnostics rather than primary evidence,
as low-dimensional projections may distort global geometry.
Accordingly, all claims in the paper are grounded in the distance-based probes reported in the
main text and appendix tables.

Each dashboard visualizes (i) a 2D projection of node embeddings,
(ii) a 3D embedding visualization,
(iii) the underlying relational graph for a representative node,
and (iv) the corresponding softmax successor probability distribution.All visualizations are generated from graphs with 60 nodes, using embeddings of dimension 128.

\subsection{Layer-Wise Geometric Properties}
\label{app:layerwise_geometry}

In addition to our analysis of the final-layer representations presented in the main text, we extract and evaluate the geometric properties of the network at every intermediate layer. Table~\ref{tab:layerwise_mega} provides the complete layer-wise profile (from the initial input embedding at L0 to the final layer at L8) across all five graph families for the single-token successor prediction task. This includes the Spearman rank correlation ($\rho$), the coefficient of determination ($R^2$), Euclidean distortion, Expansion Variance, and Local Isometry. The results are aggregated over all independent model seeds and detail the progression of the representations throughout the depth of the network.
\subsection{Ablation on Edge Directionality}
\label{app:direction_ablation}

To rule out edge directionality as a potential confounding factor in our primary results, we conducted an ablation study where all five graph topologies were constructed with strictly bidirectional (undirected) edges. Table~\ref{tab:bidirectional_all} presents the geometric emergence metrics for this fully bidirectional construct. The results demonstrate that the qualitative ordering of geometric recoverability remains completely unchanged: Watts-Strogatz achieves the strongest spatial organization, followed by 3-Regular, with Cycle and Chain trailing significantly behind, and the Star graph failing to recover a valid geometry. This confirms that our findings are driven by the intrinsic structural properties of the topologies (such as clustering and target entropy), rather than artifacts of directed versus undirected edge definitions.

\section{Architecture Independence}
\label{app:architecture_independence}

\subsection{Encoder-Decoder Transformers}
\label{app:encoder_decoder}

In the main text, our analyses focus on decoder-only Transformers to align with common architectures used in autoregressive language modeling. To verify that the observed relationship between memorization, topological symmetry, and geometric emergence is a fundamental property of the learning objective rather than an architectural artifact, we replicate our core experiments using an encoder-decoder framework. We adopt a standard encoder-decoder Transformer trained from scratch under the identical successor-prediction objective. To maintain capacity comparability with the decoder-only experiments—which utilized 8 layers—we configure the encoder-decoder model to have 4 encoder layers and 4 decoder layers, yielding an equivalent total depth. The embedding dimension is fixed to $d=128$ with 4 attention heads, and the graph size to $n=60$. As with the primarily analyzed models, the encoder-decoder configurations achieve near-perfect memorization (set accuracy $\geq 99.9\%$) across all tested graph families.

We probe the internal representations to determine whether introducing an encoder-decoder architecture alters geometric recoverability. Our findings demonstrate that the representations exhibit the precise pattern of geometry emergence observed in decoder-only models (Table~\ref{tab:geom_encdec_decoder}). In highly symmetric topologies, such as chains and cycles, the model does not develop a coherent global metric. The learned embeddings exhibit near-zero rank correlation with graph shortest-path distances and high distortion, confirming that high node interchangeability permits non-geometric memorization regardless of the encoder-decoder architecture. Under partial symmetry-breaking, such as in 3-regular graphs, the representation space shows only weak and fragile geometric alignment, characterized by inconsistent global scaling.

Conversely, under the strong symmetry-breaking of Watts--Strogatz small-world graphs, coherent and stable geometric structure robustly emerges within the model's internal representations. In the boundary case of star graphs, the geometry degenerates into the trivial, topology-dictated hub-and-spoke separation, confirming it as an imposed limit rather than a learned spatial organization. These results establish that the selective emergence of geometric structure is architecture-agnostic. The non-identifiability of metric structure in symmetric graphs is a direct mathematical consequence of how the task objective interacts with relational topology, rather than an artifact of causal masking or decoder-only processing. Consequently, the conclusion that geometric structure depends primarily on symmetry constraints applies broadly across Transformer architectures.

\subsection{The Role of Self-Attention and Non-Linearity}
\label{app:arch_ablation}

To further determine whether continuous geometry is an artifact of the Transformer's self-attention mechanism, we evaluated the Watts-Strogatz and Chain topologies across a broader spectrum of neural architectures. Chain is included alongside Watts--Strogatz specifically as a topology-level control: it lets us check whether any architecture can manufacture recoverable geometry where the topology does not support it, rather than only checking whether architectures preserve geometry where the topology already supports it. Because the successor prediction objective requires only a single input token, the attention matrix in standard Transformer models is identically the identity matrix. Consequently, a single-token Transformer functions mathematically as a deep feed-forward residual network.

As shown in Table~\ref{tab:arch_ablation}, self-attention is entirely unnecessary for the emergence of geometry on Watts--Strogatz. A deep residual network with all attention mechanisms removed (\texttt{resid\_noattn}) achieves geometric alignment comparable to the full attention-based Transformer ($\rho = 0.568 \pm 0.022$ vs. $\rho = 0.630 \pm 0.034$).

Furthermore, we find that deep non-linear architectures are not strictly required. While a parameter-free lookup table shows the weakest alignment among the architectures tested ($\rho = 0.206 \pm 0.025$) due to the lack of representational compression, introducing a simple linear projection over the embeddings (\texttt{emb\_linear}) is sufficient to bring alignment close to the full model's level ($\rho = 0.549 \pm 0.031$). This confirms that geometric emergence is driven substantially by the presence of a continuous bottleneck, rather than any specific non-linear routing mechanism.

We also ran this same five-architecture sweep on Chain, not only on Watts--Strogatz, and this is what establishes the topology-level control: every architecture, including the full attention-based Transformer, stays near $\rho \approx 0$ on Chain (ranging from $-0.012$ to $0.039$ across the five architectures), with no architecture recovering meaningfully higher alignment than any other. Taken together, the Chain and Watts--Strogatz results indicate that the presence or absence of recoverable geometry is governed by graph topology, not by architectural capacity or the presence of self-attention: no architecture change turns a symmetric, non-geometric topology into a geometric one, and no architecture change eliminates the geometry available in an asymmetric topology.

\section{Neighborhood Reconstruction and Distribution Matching}
\label{app:neighborhood_recall}

While set accuracy confirms that the model's highest-probability prediction is a valid successor, it is also important to verify that the model memorizes the full relational neighborhood rather than collapsing its predictions to a single valid edge. Because our dataset contains uniform samples of all valid $(u, v)$ edge transitions, minimizing the cross-entropy loss over the dataset mathematically requires the model to distribute probability mass across all valid successors. We empirically evaluate this by measuring the Top-$k$ Recall, where $k$ is the true node degree in the graph. As shown in Table~\ref{tab:neighborhood_recall}, the models successfully allocate their highest logits to the exact set of valid neighbors (Recall $> 0.99$), confirming that the network has internalized the complete target distribution and full relational structure.

\section{Graph Automorphisms and Task-Induced Equivalence}
\label{app:automorphisms}
In Sections 3.4 and 4.3, we discuss how symmetry-breaking reduces the space of functionally equivalent representations, allowing geometric structure to emerge. As noted in the main text, node interchangeability can arise from either formal graph automorphisms or task-induced equivalence. 
Table~\ref{tab:graph_stats} reports the exact empirical graph-theoretic statistics for our canonical $n=60$ topologies. Crucially, randomly sampled stochastic graphs often exhibit trivial automorphism groups ($|Aut(G)| = 1$) and fully unique node orbits. Because both our sampled 3-Regular and Watts-Strogatz graphs are globally asymmetric, the stronger symmetry-breaking observed in the Watts-Strogatz topology cannot be attributed to the collapse of formal global automorphisms, as they are already broken in both cases.
Instead, the symmetry-breaking that separates Watts-Strogatz from 3-Regular is driven by the collapse of functional (task-induced) symmetries. We operationalize this using Target Entropy, which quantifies the richness and overlap of the conditional successor distributions induced by the task. The Watts-Strogatz graph exhibits significantly higher local clustering and higher Target Entropy than the 3-Regular graph. It is this richness in local, overlapping transition dynamics that breaks functional symmetries, providing the necessary constraints for the cross-entropy objective to pull topologically related nodes into a continuous spatial manifold.
To explicitly isolate the effect of formal automorphisms from functional task-induced equivalence, we evaluate a Watts-Strogatz rewiring sweep in Table~\ref{tab:ws_sweep}. At $p=0.0$ (a perfect ring lattice), the graph possesses massive formal symmetry ($|Aut(G)|=120$), which creates unidentifiability and suppresses geometric emergence ($\rho=0.56$). Rewiring a small fraction of edges ($p=0.1$ and $p=0.2$) completely destroys the formal global symmetry ($|Aut(G)|=1$) while preserving the high Target Entropy and local clustering. By breaking the formal automorphism, the unidentifiability is resolved, allowing the model to recover a vastly superior geometry ($\rho > 0.63$). This confirms that the collapse of formal structural symmetry is necessary, but the geometry itself is fundamentally driven by the functional task constraints.

\section{Robustness to Scale: Normalized and Scale-Free Metrics}
\label{app:scale_free_metrics}

To ensure that the geometric structures observed in the main text are fundamentally topological and not artifacts of arbitrary volume scaling in the Transformer's hidden states, we evaluate two complementary approaches: computing scale-invariant companions to our global structural metrics, and computing scale-free local neighborhood-preservation measures.

\subsection{Scale-Invariant Structural Metrics}
\label{app:scale_invariant_metrics}

Two of the five structural metrics in the main manuscript, Expansion Variance and Local Isometry, are computed directly from raw embedding distances and therefore carry the arbitrary scale of the network's hidden states: rescaling every embedding by a constant factor changes their numerical value even though the underlying geometry is unchanged. The remaining three metrics, Spearman $\rho$, $R^2$, and Distortion, are already scale-invariant by construction, since each is a ratio- or rank-based statistic in which any global scale factor cancels; their values here are therefore unchanged from the main manuscript.

To isolate a scale-invariant counterpart to Expansion Variance and Local Isometry, we report Exp. CV, the coefficient of variation of the per-pair expansion ratio $d_E(u,v)/d_G(u,v)$, and 1-hop Norm, the local isometry statistic divided by the mean embedding distance. Both are computed as ratios of raw-scale quantities from the original embeddings, rather than by rescaling or projecting the embeddings themselves, and both cancel any global multiplicative rescaling of the representation in the same way Distortion and Spearman $\rho$ already do. This isolates the geometry induced by the training objective and graph topology from the overall magnitude of the hidden-state activations. These two columns are also reported alongside the raw metrics in every dimension-sweep, graph-size-sweep, and loss-ablation table in Appendix~\ref{app:dim_sweep}--\ref{app:graph_size}, so that the comparisons made across those tables can be checked against a scale-invariant measure directly rather than relying on Expansion Variance and Local Isometry alone.

Table~\ref{tab:normalized_metrics} reports the normalized values for each of the five graph families, confirming that the relative geometric alignment observed in the unnormalized metrics holds consistently under rigorous normalization.

\subsection{Scale-Free Neighborhood-Preservation Measures}
\label{app:neighborhood_metrics}

To further verify local structure independently of global scale, we evaluate several scale-free neighborhood-preservation measures. As shown in Table~\ref{tab:neighborhood_metrics}, we compute:
\begin{itemize}
    \item \textbf{Separation Ratio:} The ratio of the average embedding distance to a node's true graph neighbors versus the average embedding distance to non-neighbors. A ratio near $1.0$ indicates that neighbors and non-neighbors sit at indistinguishable distances in the embedding space, i.e., no local separation; values well below $1.0$ indicate that true graph neighbors are consistently embedded closer together than non-neighbors, reflecting locally coherent structure. Values above $1.0$, as with the star graph, indicate an inverted relationship in which nominal neighbors are on average embedded \emph{farther} apart than non-neighbors.
    \item \textbf{Trustworthiness \& Continuity:} Standard dimensionality-reduction diagnostics, adapted here to compare graph-defined neighborhoods against embedding-defined neighborhoods. Trustworthiness penalizes points that are close in the embedding but are not true graph neighbors (false positives), while Continuity penalizes true graph neighbors that are not embedded close together (false negatives). Both scores range from $0$ to $1$, with $1$ indicating perfect agreement between the graph and embedding neighborhoods and values near $0.5$ indicating agreement no better than chance.
\end{itemize}

Under this framing, Chain and Cycle sit close to the chance level ($\approx 0.5$) on both Trustworthiness and Continuity, and their Separation Ratios remain close to $1.0$, indicating that neighbor and non-neighbor distances are essentially indistinguishable in the embedding space -- consistent with the non-geometric, index-based memorization strategy described in Section~4. The star graph's Separation Ratio substantially exceeds $1.0$ rather than approaching it, reflecting its degenerate distance structure rather than a genuine local signal. Watts--Strogatz, and to a lesser degree 3-Regular, depart clearly from chance on all three measures, corroborating the global rank-correlation results in Table~1 with an independent, scale-free diagnostic that does not depend on absolute embedding-distance magnitudes.

\begin{figure*}[t]
    \centering
    \includegraphics[width=\textwidth]{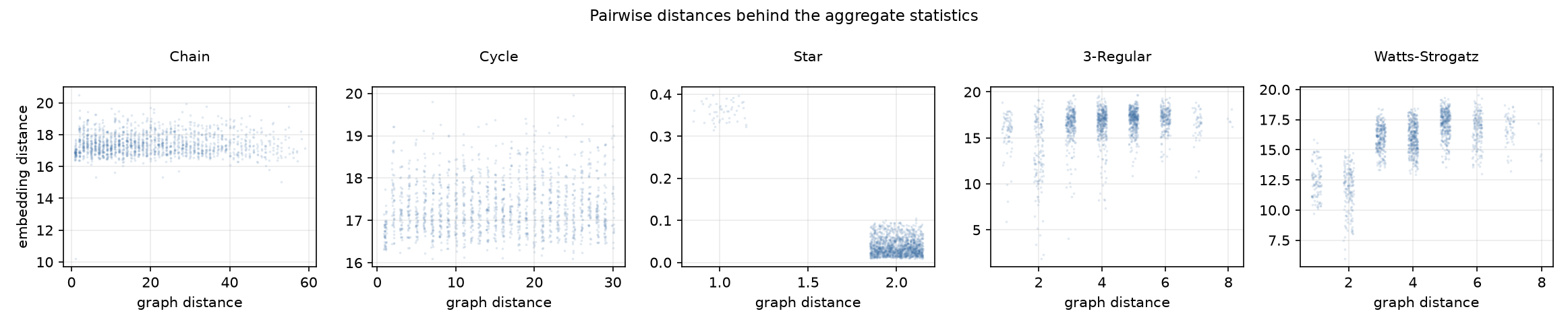}
    \caption{Pairwise distances behind the aggregate statistics. The relationship between true shortest-path graph distance ($x$-axis) and the learned $L_2$ embedding distance in the final layer ($y$-axis) for different graph topologies. Watts-Strogatz forms a clear monotonic mapping consistent with geometric emergence. Chain and Cycle show flat, unstructured distributions indicative of memorization. The Star graph exhibits a degenerate spatial embedding, characterized by a bifurcated distance distribution.}
    \label{fig:pairwise_distances}
\end{figure*}

Consistent with our normalized geometric findings, the Watts-Strogatz topology achieves the strongest local neighborhood preservation across all three measures, demonstrating robust local geometry without reliance on absolute distance scaling. To visually summarize these aggregate statistics, Figure \ref{fig:pairwise_distances} provides the pairwise distance plots for these topologies.

\begin{table*}[t]
\centering
\resizebox{\textwidth}{!}{%
\begin{tabular}{lccccccc}
\toprule
Graph & Spearman $\rho$ & $R^2$ & Distortion & Exp. Var. & Local Iso. & Exp. CV & 1-hop Norm \\
\midrule
Chain           & $0.0369 \pm 0.0428$ & $0.0020 \pm 0.0033$ & $71.28 \pm 9.71$ & $5.0873 \pm 0.1153$ & $11.1921 \pm 0.0982$ & $1.463 \pm 0.003$ & $0.887 \pm 0.005$ \\
Cycle           & $0.0534 \pm 0.0144$ & $0.0022 \pm 0.0017$ & $64.74 \pm 1.66$ & $5.0867 \pm 0.0957$ & $11.2113 \pm 0.0911$ & $1.464 \pm 0.006$ & $0.890 \pm 0.009$ \\
3-Regular       & $0.6177 \pm 0.0466$ & $0.2567 \pm 0.0522$ & $5.78  \pm 1.34$ & $10.6819 \pm 0.5754$ & $11.2883 \pm 0.4635$ & $0.368 \pm 0.025$ & $0.865 \pm 0.016$ \\
Watts--Strogatz & $0.6466 \pm 0.0521$ & $0.4701 \pm 0.0681$ & $8.12  \pm 0.50$ & $3.9376 \pm 0.9151$ & $10.1883 \pm 1.2606$ & $0.431 \pm 0.036$ & $0.677 \pm 0.063$ \\
Star            & N/A$^{*}$ & $0.000 \pm 0.000$ &$18.38 \pm 12.60$ & $3.7315 \pm 3.8629$ & $3.7698 \pm 3.9862$ & $0.807 \pm 0.798$ & $1.000 \pm 0.000$ \\
\bottomrule
\end{tabular}
}
\caption{Geometry diagnostics for $n=60$, $d=32$ using the final layer (Layer~8). Multi-seed. $^{*}$Spearman $\rho$ is undefined for Star under the directed convention (only hub--spoke pairs are reachable, so $\mathrm{Var}(d_G){=}0$). $R^2$ is well-defined in this regime: the least-squares fit reduces to the constant predictor $\hat d_E{=}\bar d_E$, giving $R^2{=}0$ exactly for any embedding, which we report rather than omit.}
\label{tab:geom_n60_d32}
\end{table*}

\begin{table*}[t]
\centering
\resizebox{\textwidth}{!}{%
\begin{tabular}{lccccccc}
\toprule
Graph & Spearman $\rho$ & $R^2$ & Distortion & Exp. Var. & Local Iso. & Exp. CV & 1-hop Norm \\
\midrule
Chain           & $0.0472 \pm 0.0249$ & $0.0019 \pm 0.0020$ & $60.59 \pm 1.50$ & $6.6635 \pm 0.1110$ & $13.0606 \pm 0.0718$ & $1.492 \pm 0.019$ & $0.930 \pm 0.026$ \\
Cycle           & $0.0541 \pm 0.0233$ & $0.0019 \pm 0.0013$ & $62.18 \pm 1.20$ & $6.7102 \pm 0.1437$ & $13.1429 \pm 0.0479$ & $1.493 \pm 0.018$ & $0.933 \pm 0.025$ \\
3-Regular       & $0.5371 \pm 0.0481$ & $0.1707 \pm 0.0413$ & $6.50  \pm 2.91$ & $15.7032 \pm 0.6754$ & $13.2684 \pm 0.5645$ & $0.392 \pm 0.026$ & $0.909 \pm 0.015$ \\
Watts--Strogatz & $0.6218 \pm 0.0253$ & $0.4162 \pm 0.0275$ & $8.77  \pm 0.95$ & $5.7175 \pm 0.2083$ & $12.6424 \pm 0.2235$ & $0.497 \pm 0.015$ & $0.820 \pm 0.024$ \\
Star            & N/A$^{*}$ & $0.000 \pm 0.000$ & $28.11 \pm 22.60$ & $10.1507 \pm 9.3842$ & $3.9635 \pm 4.5440$ & $1.133 \pm 1.062$ & $1.000 \pm 0.000$ \\
\bottomrule
\end{tabular}
}
\caption{Geometry diagnostics for $n=60$, $d=64$ using the final layer (Layer~8). Multi-seed. $^{*}$Spearman $\rho$ is undefined for Star under the directed convention (only hub--spoke pairs are reachable, so $\mathrm{Var}(d_G){=}0$). $R^2$ is well-defined in this regime: the least-squares fit reduces to the constant predictor $\hat d_E{=}\bar d_E$, giving $R^2{=}0$ exactly for any embedding, which we report rather than omit.}
\label{tab:geom_n60_d64}
\end{table*}

\begin{table*}[t]
\centering
\resizebox{\textwidth}{!}{%
\begin{tabular}{lccccccc}
\toprule
Graph & Spearman $\rho$ & $R^2$ & Distortion & Exp. Var. & Local Iso. & Exp. CV & 1-hop Norm \\
\midrule
Chain           & $0.0692 \pm 0.0195$ & $0.0023 \pm 0.0019$ & $65.98 \pm 2.01$ & $18.4257 \pm 0.1198$ & $22.0594 \pm 0.0740$ & $1.508 \pm 0.003$ & $0.958 \pm 0.004$ \\
Cycle           & $0.0670 \pm 0.0206$ & $0.0032 \pm 0.0022$ & $63.96 \pm 4.00$ & $18.4965 \pm 0.1487$ & $22.0563 \pm 0.0385$ & $1.504 \pm 0.014$ & $0.951 \pm 0.017$ \\
3-Regular       & $0.2776 \pm 0.1203$ & $0.0463 \pm 0.0318$ & $10.41 \pm 6.03$ & $24.8055 \pm 0.9689$ & $16.3252 \pm 0.8486$ & $0.415 \pm 0.027$ & $0.964 \pm 0.023$ \\
Watts--Strogatz & $0.6426 \pm 0.0294$ & $0.4539 \pm 0.0345$ & $8.81  \pm 0.76$ & $5.1551 \pm 0.2100$ & $11.9139 \pm 0.2455$ & $0.506 \pm 0.016$ & $0.835 \pm 0.027$ \\
Star            & N/A$^{*}$ & $0.000 \pm 0.000$ & $1.69 \pm 0.18$ & $0.0005 \pm 0.0005$ & $0.2687 \pm 0.0890$ & $0.095 \pm 0.025$ & $1.000 \pm 0.000$ \\
\bottomrule
\end{tabular}
}
\caption{Geometry diagnostics for $n=60$, $d=256$ using the final layer (Layer~8). Multi-seed. $^{*}$Spearman $\rho$ is undefined for Star under the directed convention (only hub--spoke pairs are reachable, so $\mathrm{Var}(d_G){=}0$). $R^2$ is well-defined in this regime: the least-squares fit reduces to the constant predictor $\hat d_E{=}\bar d_E$, giving $R^2{=}0$ exactly for any embedding, which we report rather than omit.}
\label{tab:geom_n60_d256}
\end{table*}

\begin{table*}[t]
\centering
\resizebox{\textwidth}{!}{%
\begin{tabular}{lccccccc}
\toprule
Graph & Spearman $\rho$ & $R^2$ & Distortion & Exp. Var. & Local Iso. & Exp. CV & 1-hop Norm \\
\midrule
Chain           & $0.0344 \pm 0.0187$ & $0.0011 \pm 0.0013$ & $64.30 \pm 3.04$ & $10.6744 \pm 0.0568$ & $16.7434 \pm 0.0548$ & $1.512 \pm 0.002$ & $0.958 \pm 0.003$ \\
Cycle           & $0.0484 \pm 0.0174$ & $0.0015 \pm 0.0009$ & $61.83 \pm 1.34$ & $10.7379 \pm 0.1500$ & $16.8388 \pm 0.1077$ & $1.514 \pm 0.001$ & $0.964 \pm 0.002$ \\
3-Regular       & $0.5112 \pm 0.1067$ & $0.1531 \pm 0.0622$ & $6.39  \pm 2.54$ & $22.2759 \pm 2.1109$ & $15.9790 \pm 0.4859$ & $0.391 \pm 0.030$ & $0.923 \pm 0.016$ \\
Watts--Strogatz & $0.5062 \pm 0.0234$ & $0.3263 \pm 0.0213$ & $8.90  \pm 0.61$ & $10.6871 \pm 0.3691$ & $16.8832 \pm 0.3211$ & $0.532 \pm 0.019$ & $0.872 \pm 0.020$ \\
Star            & N/A$^{*}$ & $0.000 \pm 0.000$ & $1.75 \pm 0.16$ & $0.0011 \pm 0.0006$ & $0.2716 \pm 0.0588$ & $0.116 \pm 0.017$ & $1.000 \pm 0.000$ \\
\bottomrule
\end{tabular}
}
\caption{Geometry diagnostics using uniform $1/K$ loss ($n=60$, $d=128$), final layer (Layer~8). Multi-seed. $^{*}$Spearman $\rho$ is undefined for Star under the directed convention (only hub--spoke pairs are reachable, so $\mathrm{Var}(d_G){=}0$). $R^2$ is well-defined in this regime: the least-squares fit reduces to the constant predictor $\hat d_E{=}\bar d_E$, giving $R^2{=}0$ exactly for any embedding, which we report rather than omit.}
\label{tab:geom_n60_d128_invk}
\end{table*}

\begin{table*}[t]
\centering
\resizebox{\textwidth}{!}{%
\begin{tabular}{lccccccc}
\toprule
Graph & Spearman $\rho$ & $R^2$ & Distortion & Exp. Var. & Local Iso. & Exp. CV & 1-hop Norm \\
\midrule
Chain           & $-0.0129 \pm 0.1088$ & $0.0084 \pm 0.0062$ & $15.66 \pm 1.13$ & $8.6627 \pm 0.1236$ & $10.0513 \pm 0.0864$ & $0.898 \pm 0.004$ & $0.963 \pm 0.003$ \\
Cycle           & $-0.0155 \pm 0.0976$ & $0.0133 \pm 0.0061$ & $15.56 \pm 0.45$ & $9.0157 \pm 0.1383$ & $10.4347 \pm 0.0714$ & $0.900 \pm 0.004$ & $0.992 \pm 0.007$ \\
3-Regular       & $0.2496 \pm 0.1645$ & $0.0514 \pm 0.0618$ & $8.06  \pm 3.51$ & $9.6926 \pm 1.5735$ & $9.3685  \pm 0.4022$ & $0.413 \pm 0.030$ & $0.946 \pm 0.031$ \\
Watts--Strogatz & $0.5461 \pm 0.0811$ & $0.2209 \pm 0.0745$ & $5.22  \pm 0.53$ & $9.2387 \pm 0.5573$ & $10.9466 \pm 0.2277$ & $0.493 \pm 0.011$ & $0.966 \pm 0.033$ \\
Star            & N/A$^{*}$ & $0.000 \pm 0.000$ & $6.97 \pm 7.67$ & $3.0564 \pm 6.2282$ & $8.2063 \pm 6.5771$ & $0.752 \pm 0.509$ & $1.000 \pm 0.000$ \\
\bottomrule
\end{tabular}
}
\caption{Geometry diagnostics for small graphs ($n=16$, $d=32$), final layer. Multi-seed. $^{*}$Spearman $\rho$ is undefined for Star under the directed convention (only hub--spoke pairs are reachable, so $\mathrm{Var}(d_G){=}0$). $R^2$ is well-defined in this regime: the least-squares fit reduces to the constant predictor $\hat d_E{=}\bar d_E$, giving $R^2{=}0$ exactly for any embedding, which we report rather than omit.}
\label{tab:geom_n16_d32}
\end{table*}

\begin{table*}[t]
\centering
\resizebox{\textwidth}{!}{%
\begin{tabular}{lccccccc}
\toprule
Graph & Spearman $\rho$ & $R^2$ & Distortion & Exp. Var. & Local Iso. & Exp. CV & 1-hop Norm \\
\midrule
Chain           & $0.0339 \pm 0.0199$ & $0.0013 \pm 0.0011$ & $216.04 \pm 9.39$  & $13.7068 \pm 0.4589$ & $26.8377 \pm 0.5020$ & $2.074 \pm 0.011$ & $0.909 \pm 0.009$ \\
Cycle           & $0.0339 \pm 0.0200$ & $0.0013 \pm 0.0015$ & $216.71 \pm 8.53$  & $12.9969 \pm 0.1927$ & $26.1737 \pm 0.2719$ & $2.069 \pm 0.014$ & $0.907 \pm 0.013$ \\
3-Regular       & $0.1146 \pm 0.0620$ & $0.0096 \pm 0.0106$ & $10.96 \pm 1.58$   & $17.0242 \pm 1.3439$ & $13.0955 \pm 0.2441$ & $0.448 \pm 0.021$ & $0.981 \pm 0.010$ \\
Watts--Strogatz & $0.5807 \pm 0.0330$ & $0.4182 \pm 0.0150$ & $37.22 \pm 46.70$  & $2.7632  \pm 0.0817$ & $11.6258 \pm 0.2158$ & $0.472 \pm 0.005$ & $0.791 \pm 0.015$ \\
Star            & N/A$^{*}$ & $0.000 \pm 0.000$ & $2.17 \pm 1.88$ & $0.0003 \pm 0.0006$ & $5.5672 \pm 5.9697$ & $0.263 \pm 0.105$ & $1.000 \pm 0.000$ \\
\bottomrule
\end{tabular}
}
\caption{Geometry diagnostics for large graphs ($n=150$, $d=384$), final layer. Multi-seed. $^{*}$Spearman $\rho$ is undefined for Star under the directed convention (only hub--spoke pairs are reachable, so $\mathrm{Var}(d_G){=}0$). $R^2$ is well-defined in this regime: the least-squares fit reduces to the constant predictor $\hat d_E{=}\bar d_E$, giving $R^2{=}0$ exactly for any embedding, which we report rather than omit.}
\label{tab:geom_n150_d384}
\end{table*}

\begin{table}[h!]
\centering
\setlength{\tabcolsep}{2pt}
\resizebox{\textwidth}{!}{%
\begin{tabular}{llccccccccc}
\toprule
Graph & Metric & L0 & L1 & L2 & L3 & L4 & L5 & L6 & L7 & L8 \\
\midrule
\multirow{5}{*}{Chain}
 & $\rho$ & $0.028 \pm 0.024$ & $0.011 \pm 0.026$ & $0.007 \pm 0.026$ & $0.014 \pm 0.031$ & $0.011 \pm 0.033$ & $0.007 \pm 0.032$ & $-0.005 \pm 0.024$ & $-0.019 \pm 0.028$ & $0.038 \pm 0.018$ \\
 & $R^2$ & $0.004 \pm 0.003$ & $0.001 \pm 0.001$ & $0.001 \pm 0.001$ & $0.001 \pm 0.001$ & $0.001 \pm 0.002$ & $0.001 \pm 0.001$ & $0.002 \pm 0.002$ & $0.006 \pm 0.004$ & $0.001 \pm 0.001$ \\
 & Distort. & $78.5 \pm 1.7$ & $73.7 \pm 3.6$ & $71.1 \pm 5.2$ & $69.2 \pm 6.5$ & $69.9 \pm 6.3$ & $73.3 \pm 6.9$ & $77.4 \pm 6.5$ & $81.3 \pm 6.4$ & $63.7 \pm 3.35$ \\
 & Exp. Var & $0.061 \pm 0.000$ & $0.085 \pm 0.001$ & $0.106 \pm 0.003$ & $0.129 \pm 0.005$ & $0.157 \pm 0.007$ & $0.197 \pm 0.010$ & $0.260 \pm 0.012$ & $0.358 \pm 0.015$ & $10.644 \pm 0.076$ \\
 & Loc. Iso & $1.256 \pm 0.007$ & $1.449 \pm 0.010$ & $1.608 \pm 0.021$ & $1.764 \pm 0.032$ & $1.950 \pm 0.046$ & $2.200 \pm 0.054$ & $2.548 \pm 0.058$ & $3.025 \pm 0.062$ & $16.716 \pm 0.061$ \\
\midrule
\multirow{5}{*}{Cycle}
 & $\rho$ & $0.061 \pm 0.029$ & $0.068 \pm 0.034$ & $0.073 \pm 0.037$ & $0.082 \pm 0.032$ & $0.077 \pm 0.029$ & $0.071 \pm 0.025$ & $0.063 \pm 0.025$ & $0.064 \pm 0.029$ & $0.038 \pm 0.020$ \\
 & $R^2$ & $0.003 \pm 0.003$ & $0.007 \pm 0.005$ & $0.009 \pm 0.006$ & $0.011 \pm 0.006$ & $0.009 \pm 0.005$ & $0.007 \pm 0.004$ & $0.005 \pm 0.003$ & $0.004 \pm 0.004$ & $0.001 \pm 0.001$ \\
 & Distort. & $61.6 \pm 1.4$ & $63.5 \pm 2.3$ & $64.2 \pm 2.2$ & $64.4 \pm 2.2$ & $64.5 \pm 2.3$ & $64.4 \pm 2.3$ & $64.2 \pm 2.5$ & $62.9 \pm 2.1$ & $61.9 \pm 1.25$ \\
 & Exp. Var & $0.061 \pm 0.001$ & $0.084 \pm 0.001$ & $0.105 \pm 0.002$ & $0.125 \pm 0.003$ & $0.151 \pm 0.004$ & $0.189 \pm 0.006$ & $0.248 \pm 0.008$ & $0.341 \pm 0.011$ & $10.684 \pm 0.121$ \\
 & Loc. Iso & $1.265 \pm 0.007$ & $1.451 \pm 0.008$ & $1.600 \pm 0.015$ & $1.746 \pm 0.021$ & $1.920 \pm 0.027$ & $2.159 \pm 0.033$ & $2.496 \pm 0.039$ & $2.963 \pm 0.046$ & $16.790 \pm 0.098$ \\
\midrule
\multirow{5}{*}{3-Regular}
 & $\rho$ & $0.218 \pm 0.050$ & $0.018 \pm 0.081$ & $0.064 \pm 0.096$ & $0.119 \pm 0.098$ & $0.155 \pm 0.096$ & $0.184 \pm 0.092$ & $0.206 \pm 0.076$ & $0.225 \pm 0.069$ & $0.326 \pm 0.086$ \\
 & $R^2$ & $0.025 \pm 0.019$ & $0.008 \pm 0.010$ & $0.011 \pm 0.025$ & $0.016 \pm 0.025$ & $0.025 \pm 0.027$ & $0.034 \pm 0.029$ & $0.041 \pm 0.026$ & $0.047 \pm 0.030$ & $0.053 \pm 0.034$ \\
 & Distort. & $7.6 \pm 1.3$ & $11.1 \pm 3.5$ & $9.9 \pm 3.0$ & $9.4 \pm 2.3$ & $9.2 \pm 2.0$ & $9.3 \pm 2.8$ & $9.4 \pm 3.0$ & $9.6 \pm 3.4$ & $10.41 \pm 5.61$ \\
 & Exp. Var & $0.112 \pm 0.010$ & $56.661 \pm 21.217$ & $78.196 \pm 26.792$ & $107.006 \pm 38.687$ & $145.973 \pm 47.454$ & $202.418 \pm 58.237$ & $267.273 \pm 59.625$ & $344.681 \pm 64.340$ & $21.874 \pm 1.473$ \\
 & Loc. Iso & $1.087 \pm 0.032$ & $22.208 \pm 3.836$ & $26.648 \pm 4.266$ & $31.432 \pm 5.149$ & $37.017 \pm 5.669$ & $43.721 \pm 6.240$ & $50.463 \pm 6.076$ & $57.392 \pm 6.054$ & $15.498 \pm 0.585$ \\
\midrule
\multirow{5}{*}{Watts-Strogatz}
 & $\rho$ & $0.500 \pm 0.032$ & $0.394 \pm 0.071$ & $0.400 \pm 0.078$ & $0.407 \pm 0.083$ & $0.406 \pm 0.088$ & $0.406 \pm 0.093$ & $0.403 \pm 0.094$ & $0.417 \pm 0.100$ & $0.630 \pm 0.034$ \\
 & $R^2$ & $0.298 \pm 0.036$ & $0.176 \pm 0.061$ & $0.181 \pm 0.069$ & $0.187 \pm 0.075$ & $0.185 \pm 0.080$ & $0.185 \pm 0.082$ & $0.181 \pm 0.082$ & $0.194 \pm 0.085$ & $0.428 \pm 0.039$ \\
 & Distort. & $9.0 \pm 0.7$ & $12.4 \pm 1.5$ & $12.8 \pm 2.2$ & $12.7 \pm 2.2$ & $13.0 \pm 2.4$ & $12.9 \pm 2.2$ & $13.4 \pm 2.5$ & $12.3 \pm 2.4$ & $8.42 \pm 0.46$ \\
 & Exp. Var & $0.049 \pm 0.003$ & $6.644 \pm 1.930$ & $7.867 \pm 2.371$ & $9.697 \pm 2.854$ & $12.670 \pm 3.459$ & $17.213 \pm 4.113$ & $23.861 \pm 5.547$ & $32.452 \pm 8.400$ & $5.914 \pm 0.316$ \\
 & Loc. Iso & $1.153 \pm 0.037$ & $13.079 \pm 1.828$ & $14.180 \pm 2.099$ & $15.707 \pm 2.316$ & $17.891 \pm 2.436$ & $20.776 \pm 2.323$ & $24.283 \pm 2.461$ & $28.297 \pm 3.044$ & $12.833 \pm 0.339$ \\
\midrule
\multirow{5}{*}{Star}
 & $\rho$ & N/A$^{*}$ & N/A$^{*}$ & N/A$^{*}$ & N/A$^{*}$ & N/A$^{*}$ & N/A$^{*}$ & N/A$^{*}$ & N/A$^{*}$ & N/A$^{*}$ \\
 & $R^2$ & $0.000 \pm 0.000$ & $0.000 \pm 0.000$ & $0.000 \pm 0.000$ & $0.000 \pm 0.000$ & $0.000 \pm 0.000$ & $0.000 \pm 0.000$ & $0.000 \pm 0.000$ & $0.000 \pm 0.000$ & $0.000 \pm 0.000$ \\
& Distort. & $21.6 \pm 5.2$ & $30.9 \pm 13.0$ & $33.0 \pm 14.6$ & $33.5 \pm 15.1$ & $34.7 \pm 15.4$ & $35.2 \pm 15.8$ & $35.7 \pm 15.9$ & $36.0 \pm 15.8$ & $1.54 \pm 0.39$ \\
 & Exp. Var & $0.500 \pm 0.229$ & $1.120 \pm 1.002$ & $1.275 \pm 1.167$ & $1.311 \pm 1.168$ & $1.392 \pm 1.263$ & $1.454 \pm 1.332$ & $1.474 \pm 1.343$ & $1.480 \pm 1.326$ & $0.000 \pm 0.000$ \\
 & Loc. Iso & $4.072 \pm 0.990$ & $5.686 \pm 2.663$ & $6.026 \pm 2.864$ & $6.139 \pm 2.857$ & $6.307 \pm 2.982$ & $6.428 \pm 3.073$ & $6.472 \pm 3.075$ & $6.499 \pm 3.039$ & $0.015 \pm 0.009$ \\
\bottomrule
\end{tabular}
}
\caption{Layer-wise geometric properties: all metrics, all families, all depths. Multi-seed; population and convention match Table 1 exactly at L8. $^{*}$Spearman $\rho$ is undefined for Star at every layer for the same reason as in the main table: graph distance $d_G$ does not depend on layer, so $\mathrm{Var}(d_G){=}0$ holds identically at every depth. $R^2$, by contrast, is well-defined at every layer regardless of $d_E$: the least-squares fit collapses to the constant predictor, giving $R^2{=}0$ exactly.}
\label{tab:layerwise_mega}
\end{table}

\begin{table}[h!]
\centering
\small
\begin{tabular}{lccccc}
\toprule
Graph (Bidirectional) & $\rho$ & $R^2$ & Distort. & Exp. Var & Loc. Iso \\
\midrule
Chain           & $0.135 \pm 0.054$ & $0.040 \pm 0.018$ & $69.9 \pm 4.3$ & $11.1 \pm 0.3$ & $18.2 \pm 0.3$ \\
Cycle           & $0.152 \pm 0.026$ & $0.069 \pm 0.011$ & $36.5 \pm 1.3$ & $10.8 \pm 0.3$ & $18.1 \pm 0.2$ \\
3-Regular       & $0.498 \pm 0.027$ & $0.293 \pm 0.022$ & $9.6 \pm 0.8$ & $8.0 \pm 0.5$ & $15.8 \pm 0.5$ \\
Watts--Strogatz & $0.657 \pm 0.021$ & $0.460 \pm 0.019$ & $8.4 \pm 0.4$ & $5.7 \pm 0.2$ & $12.6 \pm 0.3$ \\
Star            & N/A$^{*}$ & N/A$^{*}$ & $8.08 \pm 5.22$ & $10.7 \pm 2.5$ & $18.2 \pm 2.1$ \\
\bottomrule
\end{tabular}
\caption{Geometric emergence when every graph topology is trained bidirectionally (undirected). Multi-seed . $^{*}$Undefined for Star: only two distance classes.}
\label{tab:bidirectional_all}
\end{table}
\begin{table*}[t]
\centering
\resizebox{\textwidth}{!}{%
\begin{tabular}{lccccccc}
\toprule
Graph & Spearman $\rho$ & $R^2$ & Distortion & Exp. Var. & Local Iso. & Exp. CV & 1-hop Norm \\
\midrule
Chain           & $-0.0029 \pm 0.0186$ & $0.0003 \pm 0.0004$ & $71.20 \pm 6.69$ & $14.6820 \pm 0.2376$ & $19.9017 \pm 0.1477$ & $1.543 \pm 0.002$ & $1.000 \pm 0.003$ \\
Cycle           & $0.0067 \pm 0.0147$ & $0.0003 \pm 0.0003$ & $68.01 \pm 2.91$ & $14.5646 \pm 0.2168$ & $19.8149 \pm 0.1475$ & $1.542 \pm 0.003$ & $0.998 \pm 0.003$ \\
3-Regular       & $0.1029 \pm 0.1129$ & $0.0182 \pm 0.0208$ & $8.84 \pm 5.55$ & $47.2811 \pm 3.9282$ & $22.5743 \pm 0.5544$ & $0.420 \pm 0.028$ & $0.989 \pm 0.013$ \\
Watts--Strogatz & $0.3494 \pm 0.0391$ & $0.2557 \pm 0.0352$ & $8.80 \pm 0.56$ & $0.0215 \pm 0.0022$ & $0.7413 \pm 0.0358$ & $0.563 \pm 0.023$ & $0.916 \pm 0.014$ \\
Star            & N/A$^{*}$ & $0.000 \pm 0.000$ & $6.95 \pm 8.40$ & $1.1214 \pm 1.2001$ & $2.5940 \pm 0.2378$ & $0.608 \pm 0.485$ & $1.000 \pm 0.000$ \\
\bottomrule
\end{tabular}
}
\caption{Geometry diagnostics for encoder-decoder models ($n=60$, $d=128$), final layer (Decoder Layer 4). Multi-seed. $^{*}$Spearman $\rho$ is undefined for Star under the directed convention ($\mathrm{Var}(d_G){=}0$); $R^2$ is well-defined and equals $0$ exactly, which we report rather than omit.}
\label{tab:geom_encdec_decoder}
\end{table*}

\begin{table}[h!]
\centering
\small
\setlength{\tabcolsep}{8pt}
\begin{tabular}{lcc}
\toprule
\textbf{Architecture} & \textbf{Chain} & \textbf{Watts--Strogatz} \\
\midrule
Lookup Table (No sharing)         & $0.039 \pm 0.073$  & $0.206 \pm 0.025$ \\
Embedding + Linear                & $-0.011 \pm 0.016$ & $0.549 \pm 0.031$ \\
MLP (2 Layers)                    & $0.035 \pm 0.021$  & $0.619 \pm 0.033$ \\
ResNet (8 Layers, No Attention)   & $-0.012 \pm 0.030$ & $0.568 \pm 0.022$ \\
GPT-2 (8 Layers, Full Attention)  & $0.038 \pm 0.018$  & $0.630 \pm 0.034$ \\
\bottomrule
\end{tabular}
\caption{Architecture ablation on Chain and Watts--Strogatz ($n=60$, $d=128$). Chain remains near zero and Watts--Strogatz remains substantially above it across every architecture tested, including those without self-attention, indicating that geometric emergence is governed by graph topology rather than by architecture choice. Multi-seed.}
\label{tab:arch_ablation}
\end{table}

\begin{table}[h!]
\centering
\small
\begin{tabular}{lccc}
\toprule
\textbf{Graph Family} & \textbf{Degree ($k$)} & \textbf{Edge Precision@$k$} & \textbf{Edge Recall@$k$} \\
\midrule
Chain           & $2$ & $1.0000 \pm 0.0000$ & $1.0000 \pm 0.0000$ \\
Cycle           & $2$ & $1.0000 \pm 0.0000$ & $1.0000 \pm 0.0000$ \\
3-Regular       & $3$ & $0.9964 \pm 0.0096$ & $1.0000 \pm 0.0000$ \\
Watts--Strogatz & $4$ & $0.9993 \pm 0.0016$ & $1.0000 \pm 0.0000$ \\
\bottomrule
\end{tabular}
\caption{Full neighborhood reconstruction: precision and recall of the top-$k$ predicted successors against the $k$ ground-truth neighbors, $k = \deg(u)$. Multi-seed. Star omitted: its hub/spoke degrees are $59$ and $1$ respectively, not a single $k$ comparable to the other rows.}
\label{tab:neighborhood_recall}
\end{table}

\begin{table}[h!]
\centering
\small
\setlength{\tabcolsep}{8pt}
\begin{tabular}{lcccc}
\toprule
\textbf{Graph} & \textbf{$|Aut(G)|$} & \textbf{Node Orbits} & \textbf{Target Entropy} & \textbf{Spearman $\rho$} \\
\midrule
Chain           & $1$ & $60$ & $0.00$ & $0.038 \pm 0.018$ \\
Cycle           & $60$ & $1$ & $0.00$ & $0.038 \pm 0.020$ \\
3-Regular       & $1$ & $60$ & $0.63$ & $0.326 \pm 0.086$ \\
Watts--Strogatz & $1$ & $60$ & $1.37$ & $0.630 \pm 0.034$ \\
Star            & $59!$ & $2$ & $4.08$ & N/A$^{*}$ \\
\bottomrule
\end{tabular}
\caption{Empirical graph-theoretic properties and task-induced equivalence for $n=60$ instances at  $d=128$, under each family's own trained edge direction (directed except Watts--Strogatz). Spearman $\rho$ uses the directed distance convention, matching the main table. $|Aut(G)|$ is the automorphism group of the trained relation, not the undirected base graph. $^{*}$Undefined: no distance variance.}
\label{tab:graph_stats}
\end{table}

\begin{table}[h!]
\centering
\small
\setlength{\tabcolsep}{8pt}
\begin{tabular}{lcccc}
\toprule
\textbf{Watts--Strogatz ($p$)} & \textbf{$|Aut(G)|$} & \textbf{Node Orbits} & \textbf{Target Entropy} & \textbf{Spearman $\rho$} \\
\midrule
$p=0.00$     & $120$ & $1$ & $1.39$ & $0.562 \pm 0.103$ \\
$p=0.10$     & $1$ & $60$ & $1.38$ & $0.637 \pm 0.040$ \\
$p=0.20$     & $1$ & $60$ & $1.37$ & $0.657 \pm 0.021$ \\
\bottomrule
\end{tabular}
\caption{Isolation of formal versus functional symmetry breaking under Watts--Strogatz rewiring. The perfect ring lattice ($p=0$) has a large automorphism group; minor rewiring collapses it to trivial while barely changing target entropy, and geometric emergence strengthens accordingly. Multi-seed.}
\label{tab:ws_sweep}
\end{table}

\begin{table}[h!]
\centering
\small
\setlength{\tabcolsep}{6pt}
\begin{tabular}{lccccc}
\toprule
\textbf{Graph Family} & \textbf{Spearman $\rho$} & \textbf{$R^2$} & \textbf{Distortion} & \textbf{Exp. CV} & \textbf{1-hop Norm} \\
\midrule
Chain           & $0.038 \pm 0.018$ & $0.001 \pm 0.001$ & $63.65 \pm 3.35$ & $1.511 \pm 0.001$ & $0.958 \pm 0.002$ \\
Cycle           & $0.038 \pm 0.020$ & $0.001 \pm 0.001$ & $61.90 \pm 1.25$ & $1.514 \pm 0.001$ & $0.963 \pm 0.002$ \\
3-Regular       & $0.326 \pm 0.086$ & $0.053 \pm 0.034$ & $10.41 \pm 5.61$ & $0.419 \pm 0.030$ & $0.961 \pm 0.019$ \\
Watts--Strogatz & $0.630 \pm 0.034$ & $0.428 \pm 0.039$ & $8.42 \pm 0.46$ & $0.501 \pm 0.011$ & $0.845 \pm 0.035$ \\
Star            & N/A$^{*}$ & $0.000 \pm 0.000$  & $1.54 \pm 0.39$ & $0.076 \pm 0.042$ & $1.000$ \\
\bottomrule
\end{tabular}
\caption{Scale-invariant companions to Table~1's Expansion Variance and Local Isometry, using the same population and directed convention. Multi-seed. $^{*}$Spearman $\rho$ is undefined for Star ($\mathrm{Var}(d_G){=}0$); $R^2$ is well-defined and equals $0$ exactly under the same population as Table~1.}

\label{tab:normalized_metrics}
\end{table}
\begin{table*}[t]
\centering
\small
\begin{tabular}{lccc}
\toprule
\textbf{Graph Family} & \textbf{Separation Ratio} & \textbf{Trustworthiness} & \textbf{Continuity} \\
\midrule
Chain          & $0.956 \pm 0.002$ & $0.554 \pm 0.012$ & $0.570 \pm 0.016$ \\
Cycle          & $0.962 \pm 0.002$ & $0.557 \pm 0.015$ & $0.568 \pm 0.016$ \\
Star           & $4.796 \pm 4.423$ & $0.531 \pm 0.021$ & $0.470 \pm 0.066$ \\
3-Regular      & $0.937 \pm 0.037$ & $0.708 \pm 0.031$ & $0.741 \pm 0.017$ \\
Watts-Strogatz & $0.835 \pm 0.036$ & $0.950 \pm 0.010$ & $0.944 \pm 0.016$ \\
\bottomrule
\end{tabular}
\caption{Scale-free local neighborhood-preservation measures. Uncertainty is reported as standard deviation across multi-seed runs. The Watts-Strogatz topology achieves highly distinct local separation (low Separation Ratio) and near-perfect structural Trustworthiness/Continuity ($> 0.94$).}
\label{tab:neighborhood_metrics}
\end{table*}

% FIGURE (NO \clearpage)
\begin{figure*}[t]
    \centering
    \includegraphics[width=0.5\textwidth]{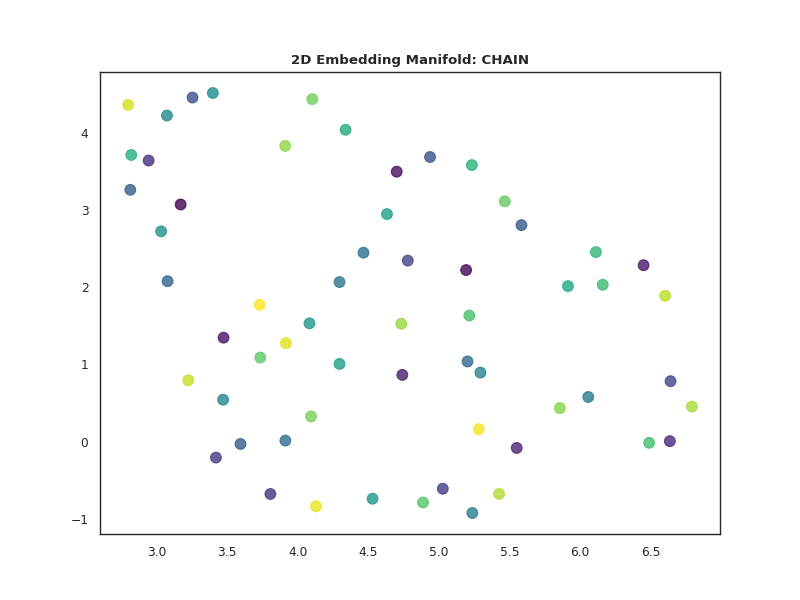}\hfill
    \includegraphics[width=0.5\textwidth]{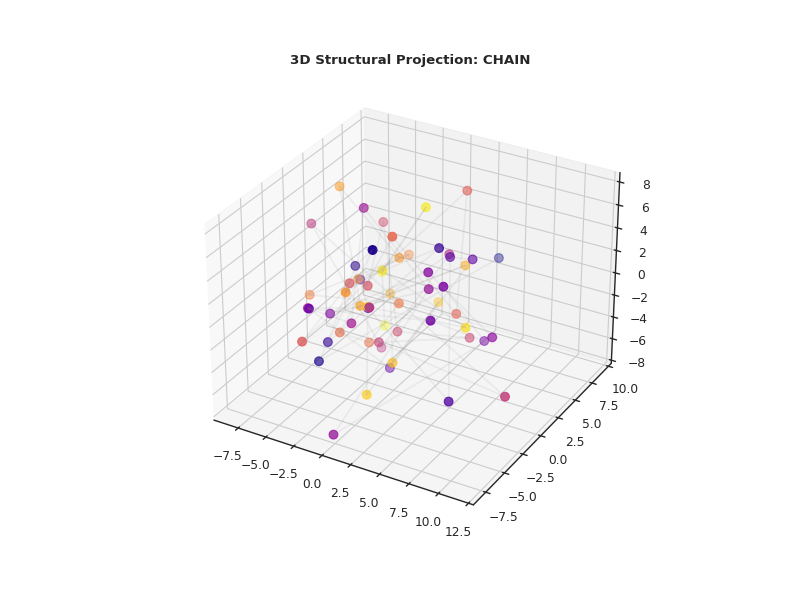}

    \vspace{0.2cm}

    \includegraphics[width=0.5\textwidth]{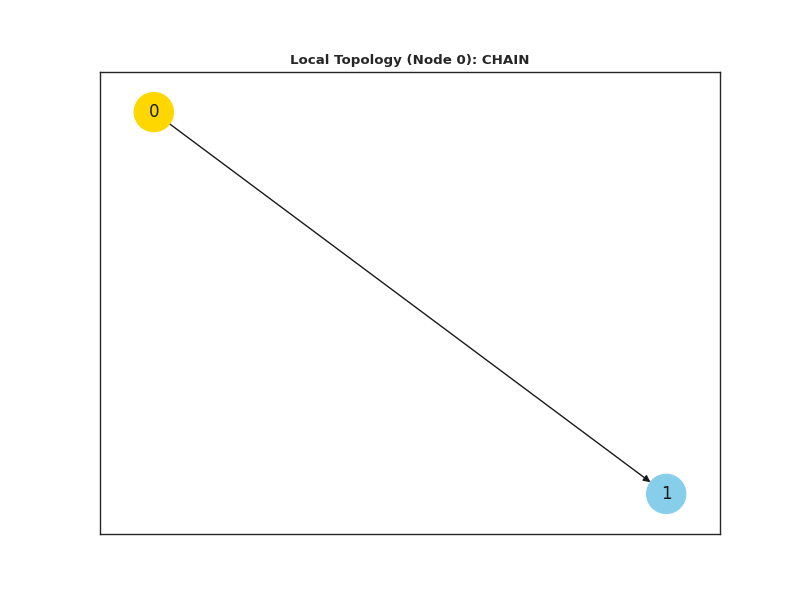}\hfill
    \includegraphics[width=0.5\textwidth]{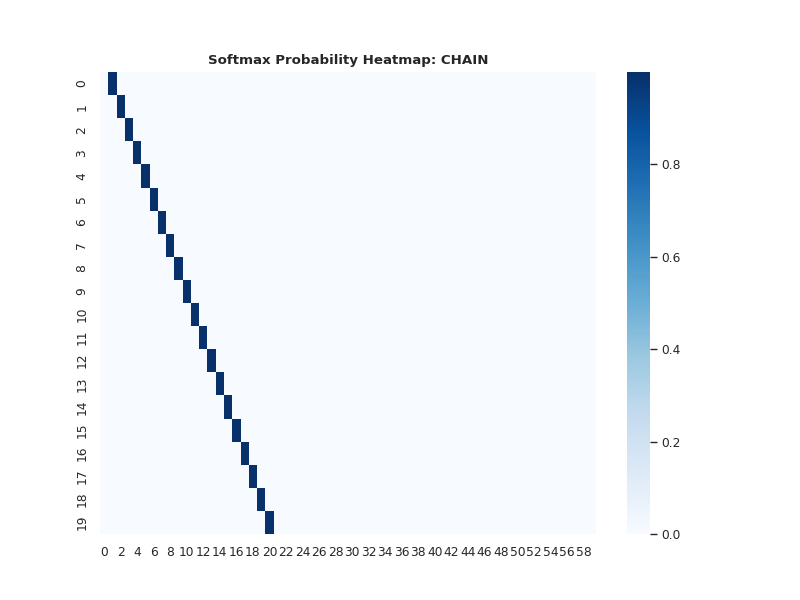}

    \caption{
    \textbf{Chain graph: memorization without geometric structure.}
    Embeddings do not exhibit a coherent global metric despite perfect task performance.
    }
    \label{fig:app_chain}
\end{figure*}

% ====================================================
% CYCLE
% ====================================================
\clearpage
\begin{figure*}[p]
    \centering
    % Row 1
    \includegraphics[width=0.5\textwidth]{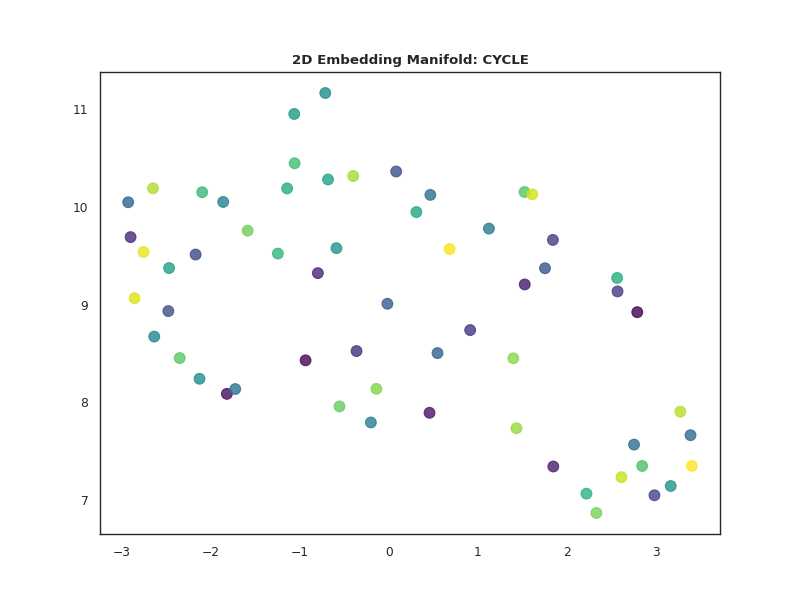}\hfill
    \includegraphics[width=0.5\textwidth]{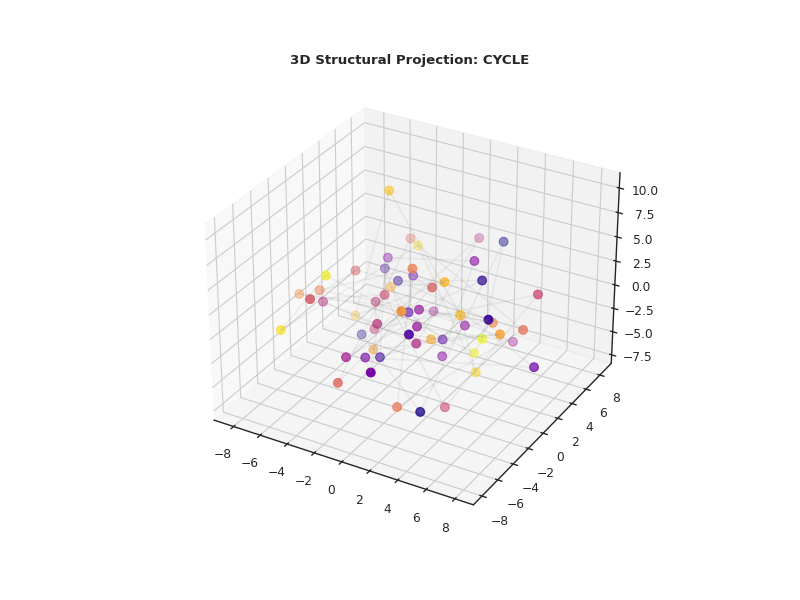}

    \vspace{0.2cm}

    % Row 2
    \includegraphics[width=0.5\textwidth]{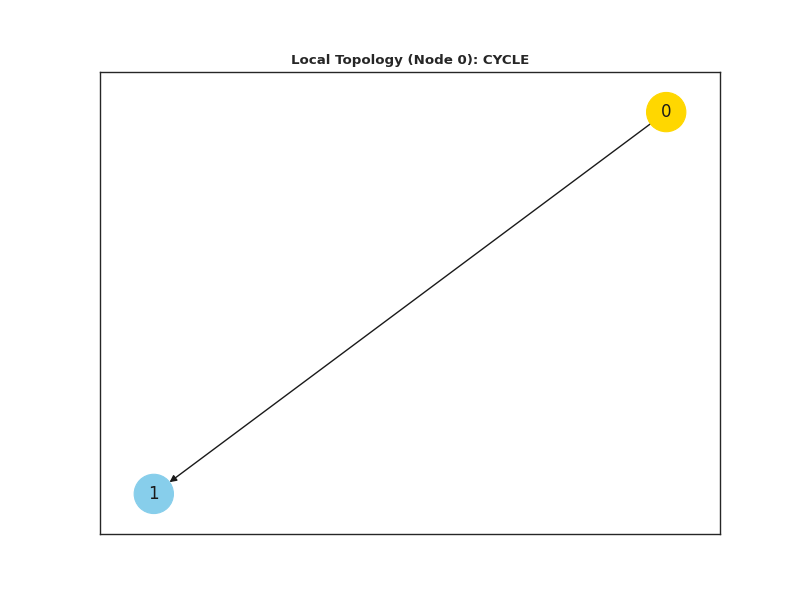}\hfill
    \includegraphics[width=0.5\textwidth]{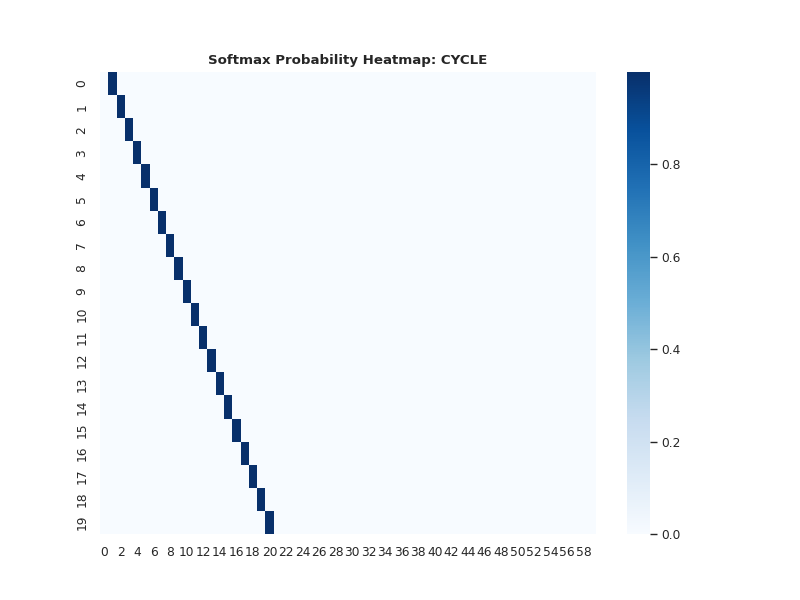}

   \caption{
\textbf{Cycle graph: symmetry-induced non-geometric memorization.}
Despite perfect memorization, embeddings do not exhibit a coherent global metric structure.
}

    \label{fig:app_cycle}
\end{figure*}

% ====================================================
% 3-REGULAR
% ====================================================
\clearpage
\begin{figure*}[p]
    \centering
    % Row 1
    \includegraphics[width=0.5\textwidth]{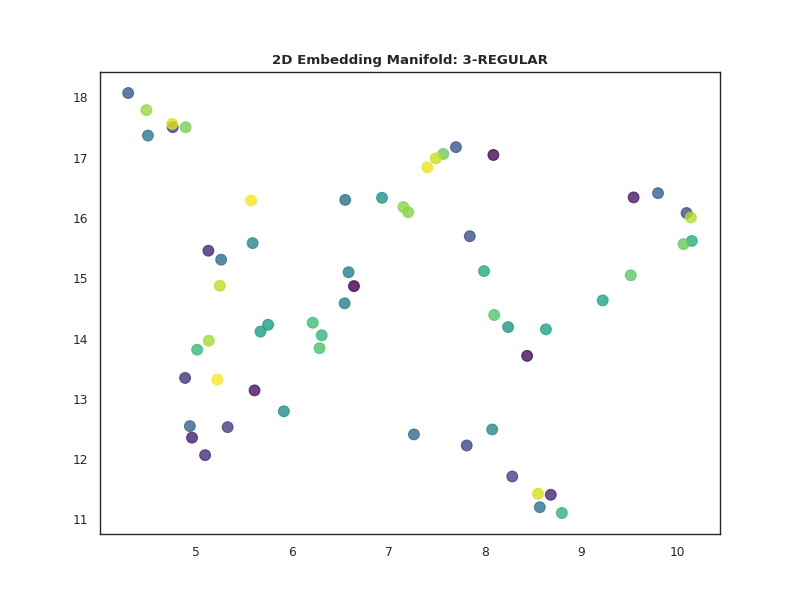}\hfill
    \includegraphics[width=0.5\textwidth]{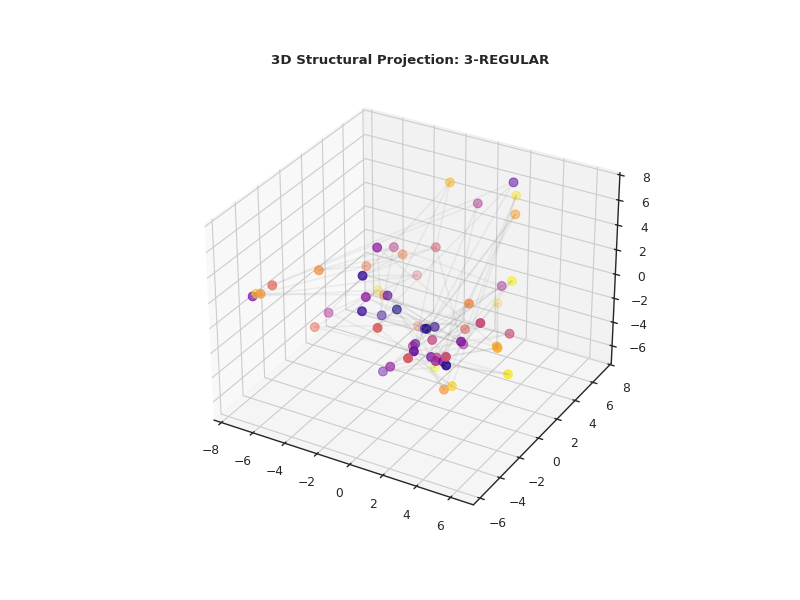}

    \vspace{0.2cm}

    % Row 2
    \includegraphics[width=0.5\textwidth]{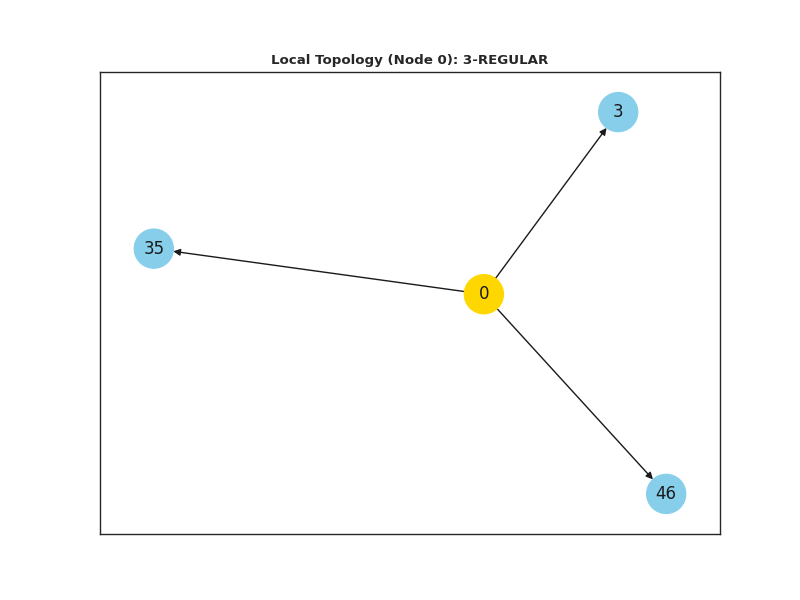}\hfill
    \includegraphics[width=0.5\textwidth]{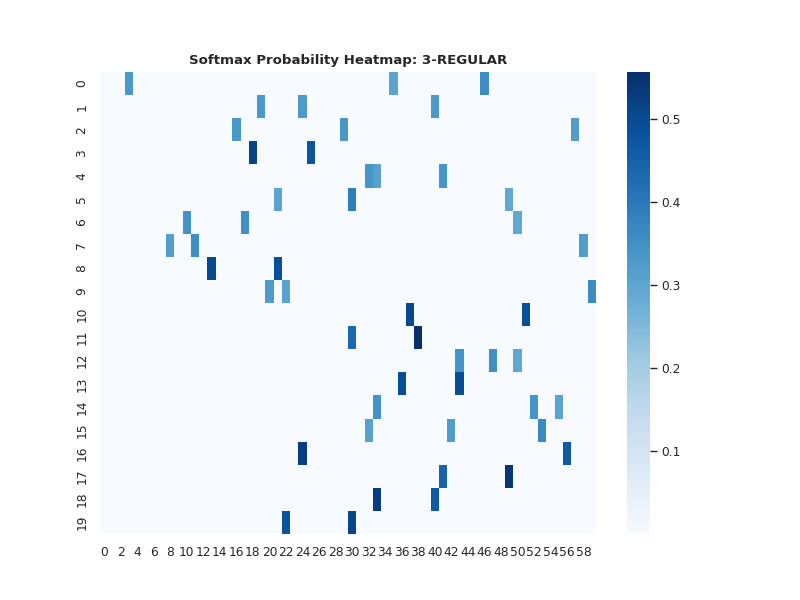}

    \caption{
    \textbf{3-Regular graph: partial asymmetry and fragile geometry.}
    Local neighborhood coherence emerges without a stable global metric,
    consistent with low-moderate rank correlation and elevated distortion.
    }
    \label{fig:app_3regular}
\end{figure*}

% ====================================================
% WATTS--STROGATZ
% ====================================================
\clearpage
\begin{figure*}[p]
    \centering
    % Row 1
    \includegraphics[width=0.5\textwidth]{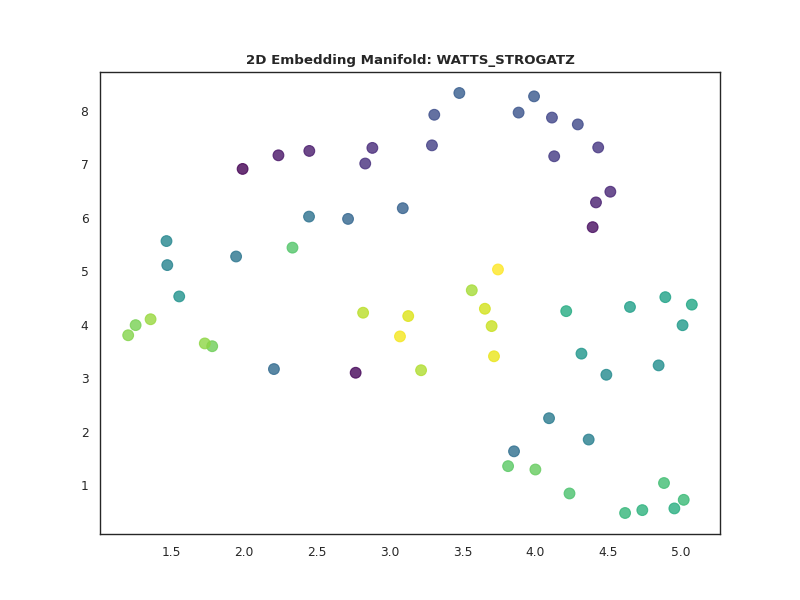}\hfill
    \includegraphics[width=0.5\textwidth]{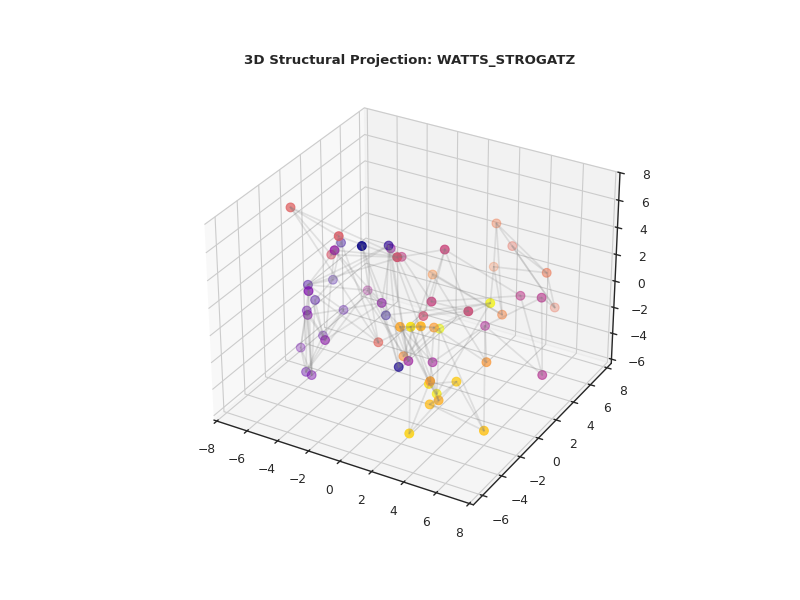}

    \vspace{0.2cm}

    % Row 2
    \includegraphics[width=0.5\textwidth]{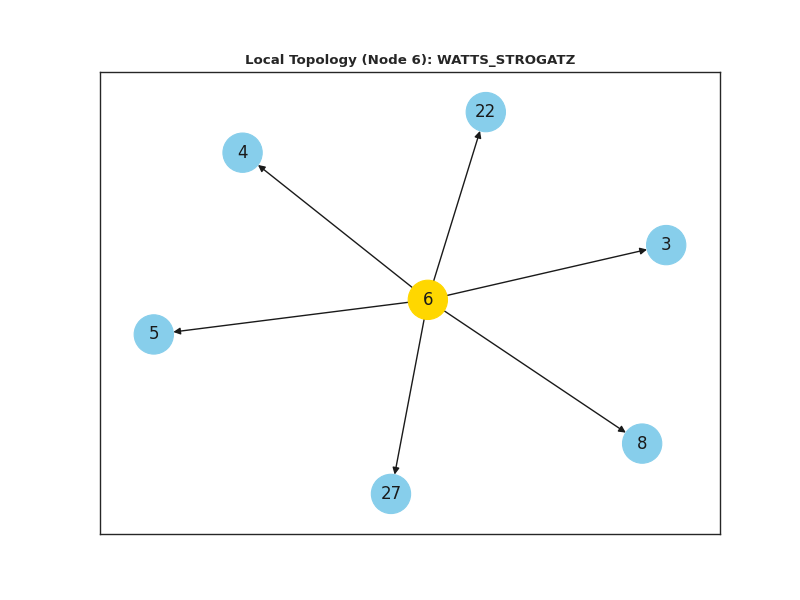}\hfill
    \includegraphics[width=0.5\textwidth]{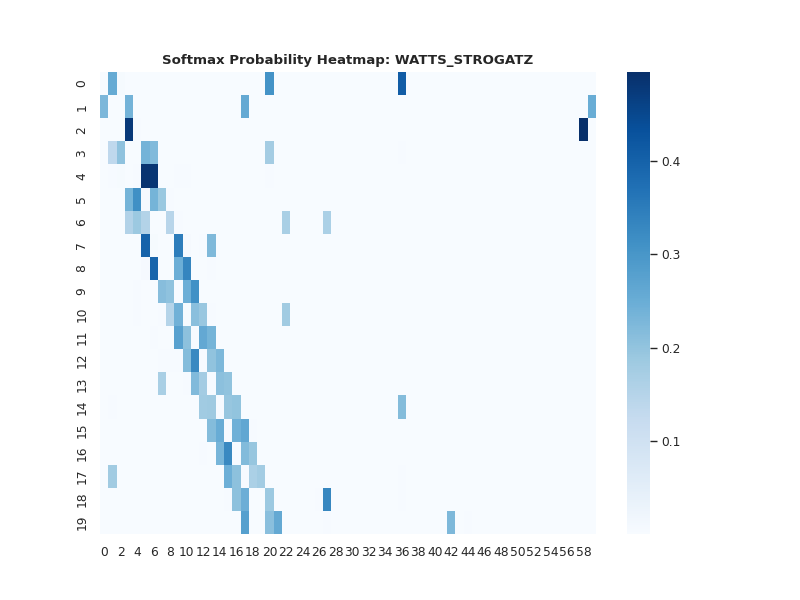}

    \caption{
    \textbf{Watts--Strogatz graph: emergent geometric organization.}
    Embeddings form a smooth low-dimensional structure that preserves graph distances,
    consistent with high rank correlation and controlled distortion.
    }
    \label{fig:app_ws}
\end{figure*}

% ====================================================
% STAR
% ====================================================
\clearpage
\begin{figure*}[p]
    \centering
    % Row 1
    \includegraphics[width=0.5\textwidth]{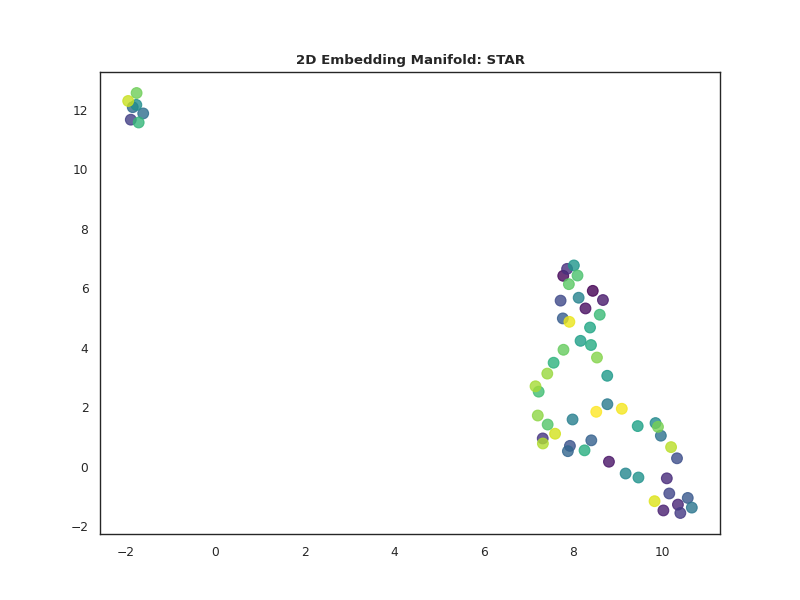}\hfill
    \includegraphics[width=0.5\textwidth]{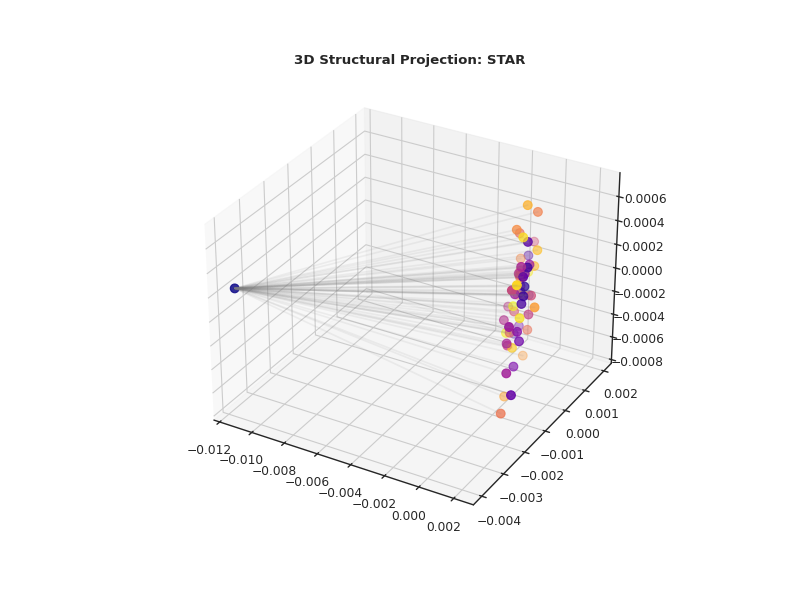}

    \vspace{0.2cm}

    % Row 2
    \includegraphics[width=0.5\textwidth]{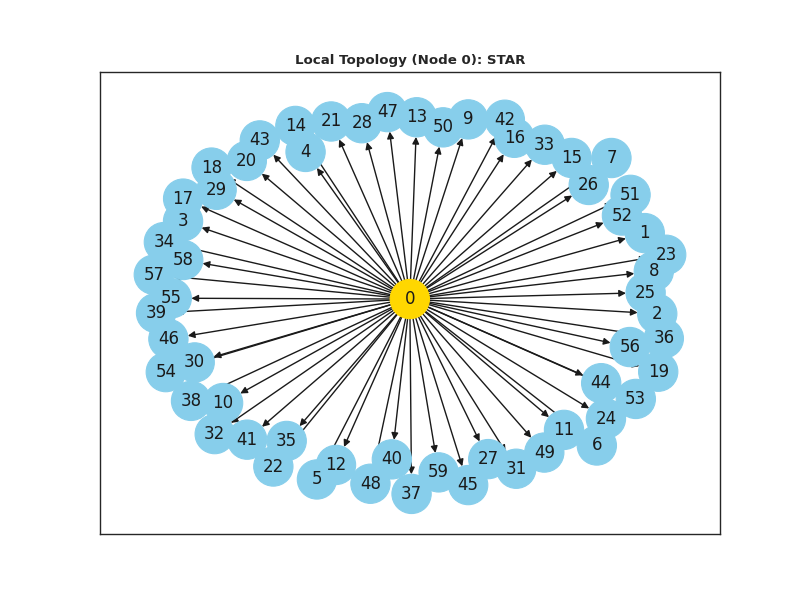}\hfill
    \includegraphics[width=0.5\textwidth]{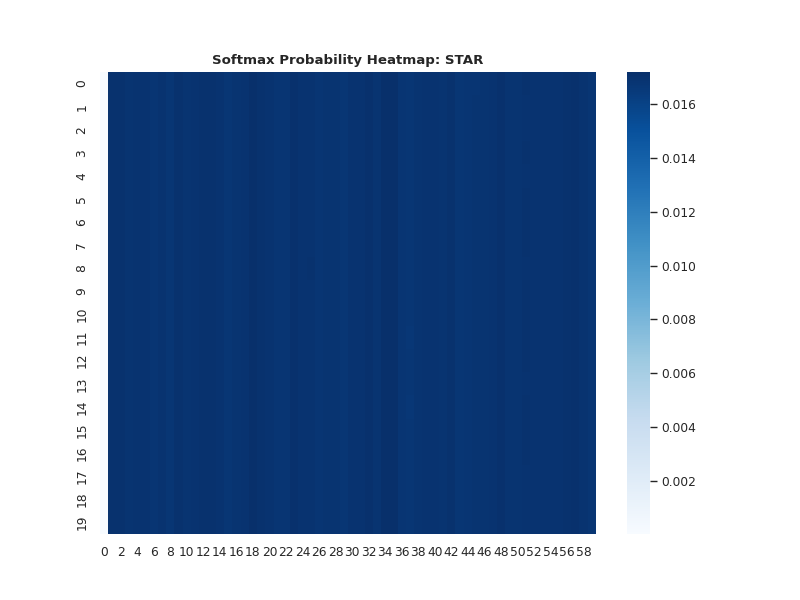}

   \caption{
\textbf{Star graph: topology-determined geometric structure.}
The hub--spoke topology yields a degenerate but task-sufficient geometric separation.
Rank-based probes are undefined due to distance collapse,
and the observed structure reflects topological degeneracy rather than recoverable metric organization.
}

    \label{fig:app_star}
\end{figure*}

\clearpage

\end{document}